\documentclass[11pt]{article}

\usepackage[final]{acl}

\usepackage{times}
\usepackage{latexsym}

\usepackage[T1]{fontenc}

\usepackage[utf8]{inputenc}

\usepackage{microtype}

\usepackage{inconsolata}

\usepackage{graphicx}

\usepackage{makecell}
\usepackage{microtype}
\usepackage{hyperref}
\usepackage{url}
\usepackage{multicol} 
\usepackage{booktabs}
\usepackage{multirow}%
\usepackage{amsmath,amssymb,amsfonts}%
\usepackage{amsthm}%
\usepackage{mathrsfs}%
\usepackage[title]{appendix}%
\usepackage[normalem]{ulem}
\usepackage{xcolor}%
\usepackage{xspace}
\usepackage{textcomp}%
\usepackage{manyfoot}%
\usepackage{booktabs}%
\usepackage{algorithm}%
\usepackage{algorithmicx}%
\usepackage{algpseudocode}%
\usepackage{listings}%
\usepackage{subcaption}
\usepackage{commath}
\usepackage{tabularray}
\usepackage{bbding}
\usepackage{pifont}
\usepackage{utfsym}
\usepackage{times}
\usepackage{wasysym}
\usepackage{fontawesome}
\usepackage[normalem]{ulem}
\usepackage{colortbl}
\usepackage{lscape}
\usepackage{enumitem}
\usepackage{fontawesome}
\usepackage{longtable}

\usepackage{amsmath}
\usepackage{thmtools}

\usepackage[most]{tcolorbox}

\declaretheoremstyle[
    spaceabove=6pt, 
    spacebelow=6pt,
    headfont=\bfseries, 
    headpunct={.}, 
    headformat={\NAME\ \NUMBER},
    bodyfont=\normalfont,
    postheadspace=0.5em
]{promptstyle}

\declaretheorem[name=Prompt, style=promptstyle]{prompt}
\declaretheorem[name=Case, style=promptstyle]{case}

\definecolor{promptColorMain}{HTML}{2E7D32} 
\definecolor{promptColorBack}{HTML}{E8F5E9} 

\definecolor{caseColorMain}{HTML}{1565C0}  
\definecolor{caseColorBack}{HTML}{E3F2FD}

\tcolorboxenvironment{prompt}{
    enhanced,              
    colback=promptColorBack,   
    colframe=promptColorMain,  
    coltitle=white,        
    fonttitle=\bfseries,   
    title={Prompt},        
    boxrule=0.8pt,         
    breakable,             
    drop shadow=black!10!white, 
    boxed title style={colback=promptColorMain}, 
    float*=t,
    width=\textwidth, 
}

\tcolorboxenvironment{case}{
    enhanced,
    colback=caseColorBack,     
    colframe=caseColorMain,    
    coltitle=white,
    fonttitle=\bfseries,
    title={Case},
    boxrule=0.8pt,
    breakable,             
    drop shadow=black!10!white,
    boxed title style={colback=caseColorMain},
    float*=t,
    width=\textwidth, 
}

\definecolor{myred}{HTML}{C00000}  
\definecolor{mygreen}{HTML}{548235}  

\newcommand{\benchname}{\textsc{AgentEHR}\xspace}
\newcommand{\agentname}{\textsc{RetroSum}\xspace}

\title{\benchname: Advancing Autonomous Clinical Decision-Making \\ via Retrospective Summarization}

\author{Yusheng Liao\thanks{Equal contribution.}$^{\spadesuit}$, 
Chuan Xuan$^{*,\spadesuit}$,
Yutong Cai$^{\spadesuit}$,
Lina Yang$^{\spadesuit}$,
Zhe Chen$^{\spadesuit}$, \\
\textbf{Yanfeng Wang$^{\spadesuit,\diamondsuit}$,
Yu~Wang\thanks{Corresponding Author}$^{,\spadesuit,\diamondsuit}$}
 \\
  $^{\spadesuit}$Shanghai Jiao Tong University \\
  $^{\diamondsuit}$Shanghai Artificial Intelligence Laboratory \\
  \texttt{\{liao20160907,xuanchuan,iautng123,alina\_yln,chenzhe2018,} \\ \texttt{wangyanfeng622,yuwangsjtu\}@sjtu.edu.cn} \\
}

\begin{document}
\maketitle

\begin{abstract} 
Large Language Models have demonstrated profound utility in the medical domain. However, their application to autonomous Electronic Health Records~(EHRs) navigation remains constrained by a reliance on curated inputs and simplified retrieval tasks. To bridge the gap between idealized experimental settings and realistic clinical environments, we present \textbf{\benchname}. This benchmark challenges agents to execute complex decision-making tasks, such as diagnosis and treatment planning, requiring long-range interactive reasoning directly within raw and high-noise databases. In tackling these tasks, we identify that existing summarization methods inevitably suffer from critical information loss and fractured reasoning continuity. To address this, we propose \textbf{\agentname}, a novel framework that unifies a retrospective summarization mechanism with an evolving experience strategy. By dynamically re-evaluating interaction history, the retrospective mechanism prevents long-context information loss and ensures unbroken logical coherence. Additionally, the evolving strategy bridges the domain gap by retrieving accumulated experience from a memory bank. Extensive empirical evaluations demonstrate that \agentname achieves performance gains of up to 29.16\% over competitive baselines, while significantly decreasing total interaction errors by up to 92.3\%. Our code and datasets are available at \url{https://github.com/BlueZeros/AgentEHR}.
\end{abstract}

\begin{figure}[!t]
    \centering
    \includegraphics[width=\linewidth]{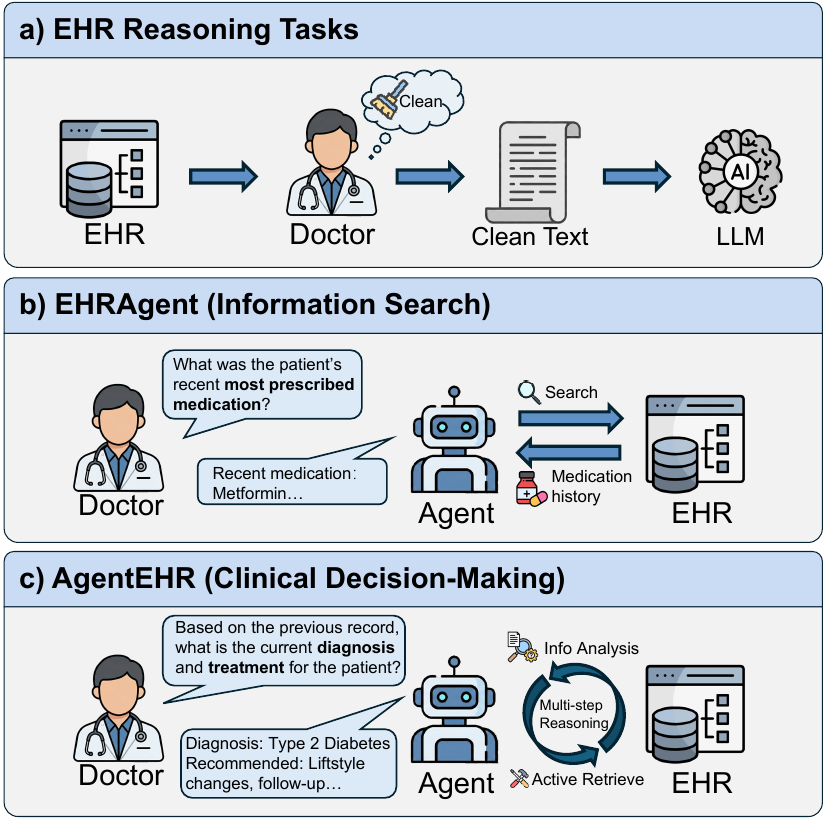}
    \caption{Comparison between the previous EHR tasks and the proposed benchmark. Unlike previous EHRAgent task retrieving factual information explicitly present in the EHR (e.g., medication history), \benchname analyzes existing information to predict future clinical decisions, like diagnosis and treatment plans.}
    \label{fig: task}
\end{figure}

\begin{table*}[t]
\centering
\resizebox{\textwidth}{!}{%
\begin{tabular}{lcccccccc}
\toprule
\multirow{2}{*}{\textbf{Methods}} & \multicolumn{5}{c}{\textbf{Agent Capacity}} & \multicolumn{3}{c}{\textbf{Agent Methods}} \\
\cmidrule(lr){2-6}\cmidrule(lr){7-9}
 & \textbf{\begin{tabular}[c]{@{}c@{}}EHR\\ Application\end{tabular}} & \textbf{\begin{tabular}[c]{@{}c@{}}Raw Data \\ Process\end{tabular}} & \textbf{\begin{tabular}[c]{@{}c@{}}Info \\ Retrieval\end{tabular}} & \textbf{\begin{tabular}[c]{@{}c@{}}Decision\\ Making\end{tabular}} & \textbf{\begin{tabular}[c]{@{}c@{}}Long\\ Context\end{tabular}} & \textbf{Summarization} & \textbf{\begin{tabular}[c]{@{}c@{}}Retrospective\\ Mechanism\end{tabular}} & \textbf{\begin{tabular}[c]{@{}c@{}}Experience\\ Accumulation\end{tabular}} \\
 \midrule
ReSum~\cite{DBLP:journals/corr/abs-2509-13313} & \textcolor{myred}{\ding{55}} & \textcolor{myred}{\ding{55}} & \textcolor{mygreen}{\ding{51}} & \textcolor{myred}{\ding{55}} & \textcolor{mygreen}{\ding{51}} & \textcolor{mygreen}{\ding{51}} & \textcolor{myred}{\ding{55}} & \textcolor{myred}{\ding{55}} \\
MEM1~\cite{zhou2025mem1} & \textcolor{myred}{\ding{55}} & \textcolor{myred}{\ding{55}} &  \textcolor{mygreen}{\ding{51}}  & \textcolor{myred}{\ding{55}} & \textcolor{mygreen}{\ding{51}} & \textcolor{mygreen}{\ding{51}} & \textcolor{myred}{\ding{55}} & \textcolor{myred}{\ding{55}} \\
Reasoning Bank~\cite{DBLP:journals/corr/abs-2509-25140} & \textcolor{myred}{\ding{55}} & \textcolor{myred}{\ding{55}} & \textcolor{myred}{\ding{55}} & \textcolor{myred}{\ding{55}} & \textcolor{myred}{\ding{55}} & \textcolor{myred}{\ding{55}} & \textcolor{myred}{\ding{55}} & \textcolor{mygreen}{\ding{51}} \\
ReflecTool~\cite{DBLP:conf/acl/LiaoJW025} & \textcolor{mygreen}{\ding{51}} & \textcolor{mygreen}{\ding{51}} & \textcolor{mygreen}{\ding{51}} & \textcolor{myred}{\ding{55}} & \textcolor{myred}{\ding{55}} & \textcolor{myred}{\ding{55}} & \textcolor{myred}{\ding{55}} & \textcolor{mygreen}{\ding{51}} \\
EHRAgent~\cite{shi-etal-2024-ehragent} & \textcolor{mygreen}{\ding{51}} & \textcolor{mygreen}{\ding{51}} & \textcolor{mygreen}{\ding{51}} & \textcolor{myred}{\ding{55}} & \textcolor{myred}{\ding{55}} & \textcolor{myred}{\ding{55}} & \textcolor{myred}{\ding{55}} & \textcolor{myred}{\ding{55}} \\
EHR-MCP~\cite{masayoshi2025ehr} & \textcolor{mygreen}{\ding{51}} & \textcolor{mygreen}{\ding{51}} & \textcolor{mygreen}{\ding{51}} & \textcolor{myred}{\ding{55}} & \textcolor{myred}{\ding{55}} & \textcolor{myred}{\ding{55}} & \textcolor{myred}{\ding{55}} & \textcolor{myred}{\ding{55}} \\
Med-Copilot~\cite{xu2025medagentgym} & \textcolor{mygreen}{\ding{51}} & \textcolor{mygreen}{\ding{51}} & \textcolor{mygreen}{\ding{51}} & \textcolor{myred}{\ding{55}} & \textcolor{myred}{\ding{55}} & \textcolor{myred}{\ding{55}} & \textcolor{myred}{\ding{55}} & \textcolor{myred}{\ding{55}} \\
EHR-R1~\cite{liao2025ehr} & \textcolor{mygreen}{\ding{51}} & \textcolor{myred}{\ding{55}} & \textcolor{myred}{\ding{55}} & \textcolor{mygreen}{\ding{51}} & \textcolor{myred}{\ding{55}} & \textcolor{myred}{\ding{55}} & \textcolor{myred}{\ding{55}} & \textcolor{myred}{\ding{55}} \\
 \midrule
\textbf{\agentname (Ours)} & \textcolor{mygreen}{\ding{51}} & \textcolor{mygreen}{\ding{51}} & \textcolor{mygreen}{\ding{51}} & \textcolor{mygreen}{\ding{51}} & \textcolor{mygreen}{\ding{51}} & \textcolor{mygreen}{\ding{51}} & \textcolor{mygreen}{\ding{51}} & \textcolor{mygreen}{\ding{51}} \\
\bottomrule
\end{tabular}%
}
\caption{Comparison of previous works and \agentname on both agent capacities and methods.}
\label{tab: comparison}
\end{table*}

\section{Introduction}

Large Language Models (LLMs) have demonstrated profound utility in the medical domain~\citep{singhal2023large, nori2023capabilities, chen2023meditron}, acting as powerful catalysts ranging from diagnostic report generation~\citep{biswas2024intelligent,jung2024enhancing} to patient communication~\citep{qiu2025evolving,qiu2025quantifying,liao2024automatic}. While these highlights underscore the potential of LLMs, effectively deploying them on the complex environment of Electronic Health Records (EHRs) remains a significant challenge.

Existing research on LLM applications in EHRs has predominantly focused on clinical decision-making tasks, such as patient risk prediction~\cite{lin2025training,fang2025toward,zhu2024realm} and intervention strategy recommendation~\cite{liao2025ehr,hager_evaluation_2024}. However, these methodologies often depend on manual curation~\cite{hegselmann2025large,hager_evaluation_2024}, creating an idealized experimental setting that sidesteps inherent challenges like data noise and long-context processing. Consequently, their applicability in realistic clinical scenarios is severely limited. Simultaneously, recent advancements have integrated agent-based systems to facilitate autonomous EHR navigation. Yet, these efforts largely confine the agent’s role to query rewriting~\cite{lee2022ehrsql,ryu2024ehr,wang2020text} or factual information retrieval~\cite{shi-etal-2024-ehragent,masayoshi2025ehr,lee2025fhir,jiang2025medagentbench}. Such approaches effectively reduce the LLM to a sophisticated search interface, failing to fully leverage the model’s advanced reasoning capabilities for complex clinical analysis.


To bridge these gaps, we introduce \textbf{\benchname}, a novel benchmark covering six diverse tasks across three real-world EHR subsets. As illustrated in Figure \ref{fig: task}, \benchname transcends the factual retrieval focus of prior work. Instead, it demands that agents actively engage in multi-step information seeking, synthesize information through clinical reasoning, and finally deliver precise decision-making predictions, such as diagnosis and treatment planning. Therefore, resolving these complex tasks necessitates extensive, multi-step interactions with the EHR system. This process inevitably accumulates a substantial amount of redundant information and results in extended interaction histories, posing a severe challenge to the model's contextual capacity and reasoning integrity.

A prevailing paradigm to navigate such long-horizon challenges involves the use of vanilla summarization techniques~\cite{DBLP:journals/corr/abs-2509-13313,zhou2025mem1} to distill salient information from redundancy and condense the elongated history. However, a fundamental misalignment arises when applying these paradigms to the clinical sphere. Unlike general long-context interaction tasks (e.g., deepresearch)~\cite{team2025tongyi,team2025mirothinker,chen2025medbrowsecomp}, EHR analysis is characterized by strong multi-turn correlations, as every piece of retrieved information is intrinsically linked to the same patient's physiological state. Consequently, standard unidirectional summarization inevitably severs latent cross-temporal connections, fracturing the reasoning continuity required for precise diagnosis.
To address this, we propose \textbf{\agentname}, which introduces a retrospective mechanism to capture dependencies for unbroken reasoning, alongside an evolving strategy to bridge the domain gap via experience accumulation. 
As highlighted in Table~\ref{tab: comparison}, by adopting the retrospective mechanism on EHR-based decision-making tasks, \agentname offers a promising strategy to navigate the intricacies of raw EHRs, successfully mitigating the loss of logical fidelity typically seen in clinical long-context reasoning applications.

Our main contributions are as follows: 
\begin{itemize} \item
\textbf{Realistic Clinical Benchmark:} We present \benchname, the first benchmark designed to bridge the gap between idealized settings and authentic medical scenarios. Unlike prior studies, \benchname establishes a new standard by challenging agents to perform multi-step reasoning within raw EHR databases to fulfill clinical decision-making tasks.

\item \textbf{Retrospective Reasoning Framework:} We propose \agentname, a simple yet effective method engineered to master the intricacies of long-context EHR reasoning tasks. By incorporating a retrospective mechanism, our method effectively captures latent correlative information and ensures reasoning continuity.

\item \textbf{Superior Empirical Performance:} Extensive experiments demonstrate that \agentname achieves remarkable performance gains of up to 29.16\% over existing baselines. This confirms \agentname's superior robustness and efficacy in handling the complex dynamics of real-world clinical decision-making.

\end{itemize}

\section{\benchname Benchmark}

To bridge the gap between current LLM capabilities and the complexities of real-world clinical applications, we introduce \benchname, a comprehensive evaluation framework designed for EHR-based interactive reasoning and clinical decision-making interaction.

\subsection{Data Construction} We construct \benchname based on two widely-used real-world EHR databases: MIMIC-IV~\cite{johnson2023mimic} and MIMIC-III~\cite{johnson2016mimic}. To rigorously evaluate the generalization and robustness of autonomous agents, we organize these data sources into three distinct experimental subsets~(details are shown in Appendix~\ref{appendix: data_curation}):

\paragraph{MIMIC-IV-Common (In-Distribution)} We stratify the MIMIC-IV dataset based on label frequency, selecting cases with the most prevalent conditions to form the \textit{Common} subset. This serves as the primary In-Distribution (ID) benchmark for assessing standard clinical reasoning capabilities.

\paragraph{MIMIC-IV-Rare (Label-Shift OOD)} Comprising the long-tail cases from MIMIC-IV, this subset introduces a significant distribution shift in the label space. It evaluates the agent's ability to handle low-prevalence diseases where parametric knowledge is often weaker.

\paragraph{MIMIC-III (Systemic-Shift OOD)} We utilize the complete MIMIC-III dataset to represent a more challenging systemic shift. Unlike the label-only shift in the Rare subset, MIMIC-III presents fundamental differences in table schema and information density compared to MIMIC-IV. This setting provides the most comprehensive measure of agent methods to heterogeneous EHR environments.

\subsection{Clinical Tasks} 
\benchname encompasses six core clinical tasks: \textit{Diagnoses}, \textit{Labevents}, \textit{Microbiology}, \textit{Prescriptions}, \textit{Procedures}, and \textit{Transfers}. Collectively, these tasks represent the entire lifecycle of patient hospitalization, spanning the critical phases of diagnosis, laboratory examination, and therapeutic intervention.

\subsection{Toolbox MCP Server} 
To enable the Agent system to navigate the complex EHR environment efficiently, we design a comprehensive Toolbox hosted on a Model Context Protocol (MCP) Server. This architecture provides a standardized interface for the agent to access over 19 specialized tools, ensuring robust and scalable interaction with the underlying database. The toolbox equips the model with diverse retrieval mechanisms, including temporal filtering, keyword search, fuzzy matching, and direct SQL execution. Detailed specifications of the toolbox can be found in Appendix~\ref{appendix: toolbox}.
\begin{figure*}[!t]
    \centering
    \includegraphics[width=\linewidth]{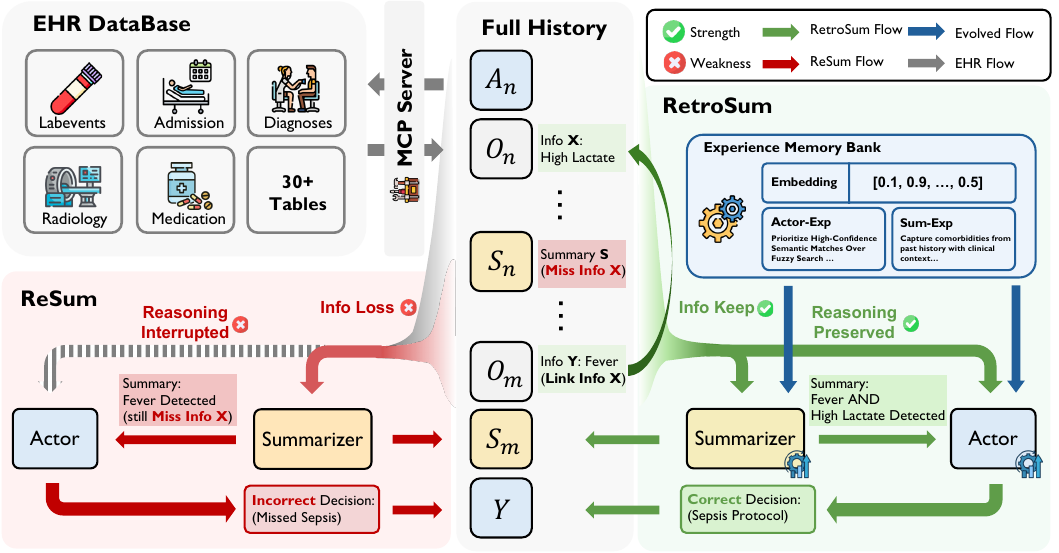}
    \caption{Overview of \agentname. \agentname~(right) addresses critical information loss and reasoning interruptions inherent in unidirectional methods like ReSum (left). By incorporating a retrospective mechanism to re-evaluate full interaction histories and an evolving mechanism to retrieve specialized strategies from memory, \agentname ensures robust long-horizon clinical reasoning and correct decisions.}
    \label{fig: overview}
\end{figure*}

\section{Methods}
In this section, we first formally define the clinical agent task. We then introduce \agentname, a novel framework designed to mitigate the information loss and reasoning fragmentation inherent in standard incremental summarization approaches (overview is shown in Figure~\ref{fig: overview}). Finally, we describe the Evolving Optimization strategy. 

\subsection{Task Formulation} 
The clinical agent task requires an agent system to interact with a database to answer complex medical queries. Each query instance is composed of $\mathcal{X}=\{p, t, I\}$, where $p$ is the patient identifier, $t$ is the reference timestamp, and $I$ is a clinical instruction. 

The interaction proceeds in discrete steps $i\in\{1, \dots, K\}$. At each step $i$, the agent observes the current state, which encapsulates the query $\mathcal{X}$ and the interaction history $H_{i-1} = \{(a_1, o_1), \dots, (a_{i-1}, o_{i-1})\}$. Based on this context, the agent generates an action $a_i$ according to its own policy $\pi_\theta$:
\begin{equation}
    a_i \sim \pi_{\theta}(a_i | H_{i-1}, \mathcal{X})
\end{equation}
Upon executing action $a_i$, the environment returns an observation $o_i$, which is derived from the EHR database $\mathcal{E}$:
\begin{equation}
    o_i = \mathcal{E}(a_i)
\end{equation}
The tuple $(a_i, o_i)$ is then appended to the history $H_i=H_{i-1}\cup(a_i, o_i)$. This process repeats until the agent issues a termination action and produces a final answer set $Y = \{y_1, y_2, \dots, y_m\}$, consisting of a list of items (e.g., specific diagnosis codes or medication names) that satisfy the clinical instruction.

\subsection{\agentname} 
\label{section: retrosum}
Navigating real-world EHRs requires the agent to manage extensive context while identifying complex dependencies across heterogeneous tables. Previous methods, such as ReSum~\cite{DBLP:journals/corr/abs-2509-13313}, typically employ unidirectional summarization, where the history context is compressed incrementally. However, we identify two critical limitations in this paradigm for clinical tasks: (1) \textbf{Loss of Latent Correlations}: EHR data is inherently interconnected. Information retrieved in early turns may initially appear irrelevant but becomes crucial after observing a later result. Unidirectional approaches, which rely on local information summarization, often discard these latent factors before their relevance is established via cross-table correlations. (2)
\textbf{Disruption of Reasoning Logic}: Forcing the agent to rely exclusively on a highly compressed summary disrupts the continuity of multi-turn reasoning logic. The abstraction process often strips away the precise syntactical and numerical details required for the agent to deduce the next logical step. To address these challenges, we propose \textbf{\agentname}, which introduces a retrospective mechanism. Specifically, our framework operates through two synergistic phases:

\paragraph{Retrospective Summarization} We define~a summarization window size $w$. The summarization process is triggered only at step $j$, where $j \equiv0 \pmod w$. At these intervals, the interaction history is conceptually partitioned into the distant history $H_\text{dist}=\{(a_1, o_1), \dots, (a_{j-w}, o_{j-w})\}$ and the recent window $H_\text{rec}=\{(a_{j-w+1}, o_{j-w+1}), \dots, (a_{j}, o_{j})\}$. The Summarizer module $\mathcal{M}$, which shares parameters $\theta$ with the agent's policy, generates an updated summary $S_j$ by retrospectively analyzing the entire trajectory:
\begin{equation}
S_j = \mathcal{M}_{\theta}(H_\text{rec}, S_{j-w}, H_\text{dist}, \mathcal{X})
\end{equation}
By conditioning on both $H_{dist}$ and $H_{rec}$, the Summarizer can re-evaluate the importance of past events in light of the most recent findings, effectively capturing latent correlations that were previously ambiguous.

\paragraph{Retrospective Inference} To prevent the disruption of reasoning logic, we design the actor to operate on a history-aware context. Unlike prior methods that replace the history with a summary, \agentname augments the full interaction history with the latest retrospective summary. For any step $i$, the entire context $\hat{H}_i$ is updated as:
\begin{equation}\small
\hat{H}_i = 
\begin{cases} 
\hat{H}_{i-1} \setminus \{S_{i-w}\} \cup \{(a_i, o_i), S_i\}, &\text{if } i \equiv 0\pmod w \\
\hat{H}_{i-1} \cup \{(a_i, o_i)\}, &\text{otherwise}
\end{cases}
\end{equation}
where $S_{i-w}$ is the most recent summary. The policy $\pi_{\theta}$ then generates the next action based on this augmented view:
\begin{equation}
a_i \sim \pi_{\theta}(a_i | \hat{H}_{i-1}, \mathcal{X})
\end{equation}
This formulation ensures that the actor retains access to the complete raw history $H_{i-1}$, preserving the integrity of the multi-turn reasoning chain, while $S$ serves as a high-level cognitive map to guide the model's attention toward clinically significant patterns extracted from previous phases.

\subsection{Evolving Strategy}
General-purpose LLMs often lack the specific clinical intuition required to distinguish subtle signals in EHRs or navigate complex table schemas. To bridge this gap, we introduce the evolving strategy for \agentname. Rather than updating model parameters, this phase enables the agent to crystallize successful strategies into an external memory bank, allowing it to learn from past trials. 

\paragraph{Experience Generation} For each training instance, the agent generates a full interaction trajectory $H_K$, a final retrospective summary $S_\text{final}$, and a predicted answer set $Y$. By comparing these outputs against the ground truth $Y^*$, we derive domain-specific experiences using the reflection module $\mathcal{R}_{\theta}$, which shares parameters with the agent's policy:
\begin{align}
E_\text{act} &= \mathcal{R}_{\theta}(H_K, Y, Y^*) \\
E_\text{sum} &= \mathcal{R}_{\theta}(H_K, S_\text{final}, Y, Y^*)
\end{align}
Here, $\mathcal{R}_{\theta}$ is prompted to extract procedural heuristics for the Actor (e.g., optimal tool selection) and information salience guidelines for the Summarizer (e.g., critical evidence retained vs. noise compressed).

\paragraph{Memory Construction} We organize these insights into an Experience Memory Bank $\mathcal{B}$. Each entry is stored as a triplet:
\begin{equation}
\mathcal{B} = \{(e, E_\text{act}, E_\text{sum})_i\}_{i=1}^M
\end{equation}
where $M$ is the size of the memory and $e$ is the representation of the EHR. $e$ is obtained by encoding the most recent clinical events from each patient's EHR tables using a pre-trained encoder. This ensures that the retrieved experience is contextually aligned with the patient's current status. 

\paragraph{Evolved Inference} During inference, for a new patient context, we compute its embedding and retrieve the most similar memory items from $\mathcal{B}$. These retrieved experiences are injected into the framework to guide both modules:
\begin{align}
S_i = \mathcal{M}_{\theta}&(H_\text{rec}, S_{i-w}, H_\text{dist}, \mathcal{X}, {E}_\text{sum}) \\ 
a_{i+1} &\sim \pi_{\theta}(a_{i+1} | \hat{H}_i, \mathcal{X}, E_\text{act})
\end{align}By explicitly conditioning on the experience, the summarizer improves its ability to filter noise, and the actor adopts proven reasoning strategies, thereby enhancing overall robustness.

\begin{table*}[t!]
\centering
\resizebox{\textwidth}{!}{%
\begin{tabular}{clcccccccc}
\toprule
\multicolumn{1}{l}{\textbf{Models}} & \textbf{Methods} & \textbf{Evolved} & \textbf{Diagnoses} & \textbf{Labevents} & \textbf{Microbiology} & \textbf{Prescriptions} & \textbf{Procedures} & \textbf{Transfers} & \textbf{Avg.} \\
\midrule
 & ReAct & \ding{55} & 0.0955 & \textbf{0.1295} & 0.1274 & 0.0756 & 0.2472 & 0.2772 & 0.1587 \\
 & Reflexion & \ding{55} & 0.1216 & 0.0983 & 0.0989 & 0.0901 & 0.2273 & 0.2292 & 0.1442 \\
 & ReSum & \ding{55} & 0.1753 & 0.0881 & 0.1166 & 0.1000 & 0.2717 & 0.2320 & 0.1639 \\
 & ReflecTool (CS) & \ding{51} & 0.1647 & 0.0821 & 0.1265 & 0.0614 & 0.2572 & 0.2589 & 0.1585 \\
 & ReflecTool (IR) & \ding{51} & 0.1046 & 0.0893 & 0.1360 & 0.0713 & 0.2428 & \uline{0.2882} & 0.1554 \\
 & ReasoningBank & \ding{51} & 0.0385 & \uline{0.1275} & 0.1250 & 0.0448 & 0.2931 & 0.2103 & 0.1399 \\
 & \cellcolor[HTML]{E1EAFF}\textbf{\agentname (Ours)} & \cellcolor[HTML]{E1EAFF}\ding{55} & \cellcolor[HTML]{E1EAFF}\uline{0.2368}$^\ddagger$ & \cellcolor[HTML]{E1EAFF}0.0977$^\dagger$ & \cellcolor[HTML]{E1EAFF}\textbf{0.1584}$^\ddagger$ & \cellcolor[HTML]{E1EAFF}\uline{0.1039} & \cellcolor[HTML]{E1EAFF}\uline{0.2983}$^\dagger$ & \cellcolor[HTML]{E1EAFF}0.2791$^\ddagger$ & \cellcolor[HTML]{E1EAFF}\uline{0.1957}$^\ddagger$ \\
\multirow{-8}{*}{\textit{Qwen3-30B-A3B}} & \cellcolor[HTML]{E1EAFF}\textbf{\agentname (Ours)} & \cellcolor[HTML]{E1EAFF}\ding{51} & \cellcolor[HTML]{E1EAFF}\textbf{0.2514}$^\ddagger$ & \cellcolor[HTML]{E1EAFF}0.1092$^\ddagger$ & \cellcolor[HTML]{E1EAFF}\uline{0.1568}$^\ddagger$ & \cellcolor[HTML]{E1EAFF}\textbf{0.1289}$^\ddagger$ & \cellcolor[HTML]{E1EAFF}\textbf{0.3068}$^\ddagger$ & \cellcolor[HTML]{E1EAFF}\textbf{0.3171}$^\dagger$ & \cellcolor[HTML]{E1EAFF}\textbf{0.2117}$^\ddagger$ \\
\midrule
 & ReAct & \ding{55} & 0.1836 & 0.0793 & \textbf{0.1895} & 0.1075 & 0.2857 & 0.2346 & 0.1800 \\
 & Reflexion & \ding{55} & 0.2323 & 0.0887 & 0.1585 & 0.0818 & 0.3098 & 0.1599 & 0.1718 \\
 & ReSum & \ding{55} & 0.2161 & 0.1083 & 0.1597 & \uline{0.1301} & 0.3503 & 0.2317 & 0.1994 \\
 & ReflecTool (CS) & \ding{51} & 0.2088 & 0.0894 & 0.1579 & 0.1059 & 0.3066 & 0.1985 & 0.1779 \\
 & ReflecTool (IR) & \ding{51} & 0.1992 & 0.0822 & 0.1389 & 0.0910 & 0.2490 & 0.2390 & 0.1665 \\
 & ReasoningBank & \ding{51} & 0.2519 & \textbf{0.1998} & 0.1726 & 0.1267 & \textbf{0.3743} & \textbf{0.2793} & \uline{0.2341} \\
 & \cellcolor[HTML]{E1EAFF}\textbf{\agentname (Ours)} & \cellcolor[HTML]{E1EAFF}\ding{55} & \cellcolor[HTML]{E1EAFF}\uline{0.2519}$^\ddagger$ & \cellcolor[HTML]{E1EAFF}0.1389$^\ddagger$ & \cellcolor[HTML]{E1EAFF}\uline{0.1881}$^\ddagger$ & \cellcolor[HTML]{E1EAFF}0.1285 & \cellcolor[HTML]{E1EAFF}0.3473 & \cellcolor[HTML]{E1EAFF}0.2217 & \cellcolor[HTML]{E1EAFF}0.2127$^\dagger$ \\
\multirow{-8}{*}{\textit{Qwen3-Next-80B-A3B}} & \cellcolor[HTML]{E1EAFF}\textbf{\agentname (Ours)} & \cellcolor[HTML]{E1EAFF}\ding{51} & \cellcolor[HTML]{E1EAFF}\textbf{0.3148}$^\ddagger$ & \cellcolor[HTML]{E1EAFF}\uline{0.1990}$^\ddagger$ & \cellcolor[HTML]{E1EAFF}0.1830$^\ddagger$ & \cellcolor[HTML]{E1EAFF}\textbf{0.1429}$^\dagger$ & \cellcolor[HTML]{E1EAFF}\uline{0.3578} & \cellcolor[HTML]{E1EAFF}\uline{0.2573}$^\dagger$ & \cellcolor[HTML]{E1EAFF}\textbf{0.2425}$^\ddagger$ \\
\midrule
 & ReAct & \ding{55} & 0.1671 & 0.1172 & 0.1545 & 0.0670 & 0.2739 & \uline{0.1413} & 0.1535 \\
 & ReSum & \ding{55} & 0.2804 & 0.1009 & \uline{0.1570} & 0.0728 & 0.2907 & 0.0803 & 0.1637 \\
 & \cellcolor[HTML]{E1EAFF}\textbf{\agentname (Ours)} & \cellcolor[HTML]{E1EAFF}\ding{55} & \cellcolor[HTML]{E1EAFF}\uline{0.2842} & \cellcolor[HTML]{E1EAFF}\uline{0.1328}$^\ddagger$ & \cellcolor[HTML]{E1EAFF}0.1568 & \cellcolor[HTML]{E1EAFF}\uline{0.0953}$^\ddagger$ & \cellcolor[HTML]{E1EAFF}\uline{0.3092} & \cellcolor[HTML]{E1EAFF}0.0988$^\dagger$ & \cellcolor[HTML]{E1EAFF}\uline{0.1795}$^\dagger$ \\
\multirow{-4}{*}{\textit{Qwen3-235B-A22B*}} & \cellcolor[HTML]{E1EAFF}\textbf{\agentname (Ours)} & \cellcolor[HTML]{E1EAFF}\ding{51} & \cellcolor[HTML]{E1EAFF}\textbf{0.3700}$^\ddagger$ & \cellcolor[HTML]{E1EAFF}\textbf{0.1951}$^\ddagger$ & \cellcolor[HTML]{E1EAFF}\textbf{0.1791}$^\ddagger$ & \cellcolor[HTML]{E1EAFF}\textbf{0.1346}$^\ddagger$ & \cellcolor[HTML]{E1EAFF}\textbf{0.3556}$^\ddagger$ & \cellcolor[HTML]{E1EAFF}\textbf{0.2342}$^\ddagger$ & \cellcolor[HTML]{E1EAFF}\textbf{0.2448}$^\ddagger$ \\
\midrule
 & ReAct & \ding{55} & \textbf{0.4270} & 0.2151 & 0.1867 & 0.0764 & \textbf{0.2693} & 0.2053 & 0.2300 \\
 & ReSum & \ding{55} & 0.4024 & 0.2181 & 0.2090 & 0.0752 & \uline{0.2668} & 0.1993 & 0.2285 \\
 & \cellcolor[HTML]{E1EAFF}\textbf{\agentname (Ours)} & \cellcolor[HTML]{E1EAFF}\ding{55} & \cellcolor[HTML]{E1EAFF}0.4023 & \cellcolor[HTML]{E1EAFF}\uline{0.2569}$^\ddagger$ & \cellcolor[HTML]{E1EAFF}\textbf{0.2146} & \cellcolor[HTML]{E1EAFF}\uline{0.1128}$^\ddagger$ & \cellcolor[HTML]{E1EAFF}0.2614 & \cellcolor[HTML]{E1EAFF}\uline{0.2071} & \cellcolor[HTML]{E1EAFF}\uline{0.2425}$^\dagger$ \\
\multirow{-4}{*}{\textit{GPT-5-mini}} & \cellcolor[HTML]{E1EAFF}\textbf{\agentname (Ours)} & \cellcolor[HTML]{E1EAFF}\ding{51} & \cellcolor[HTML]{E1EAFF}\uline{0.4127}$^\dagger$ & \cellcolor[HTML]{E1EAFF}\textbf{0.2989}$^\ddagger$ & \cellcolor[HTML]{E1EAFF}\uline{0.2097} & \cellcolor[HTML]{E1EAFF}\textbf{0.1536}$^\ddagger$ & \cellcolor[HTML]{E1EAFF}0.2603 & \cellcolor[HTML]{E1EAFF}\textbf{0.2327}$^\ddagger$ & \cellcolor[HTML]{E1EAFF}\textbf{0.2613}$^\ddagger$ \\
\midrule
 & ReAct & \ding{55} & 0.3841 & 0.1856 & 0.2032 & 0.1688 & \textbf{0.3762} & 0.1826 & \cellcolor[HTML]{FFFFFF}0.2501 \\
 & ReSum & \ding{55} & 0.3822 & 0.0882 & 0.2024 & 0.1618 & 0.3186 & 0.1888 & \cellcolor[HTML]{FFFFFF}0.2237 \\
 & \cellcolor[HTML]{E1EAFF}\textbf{\agentname (Ours)} & \cellcolor[HTML]{E1EAFF}\ding{55} & \cellcolor[HTML]{E1EAFF}\uline{0.4270}$^\ddagger$ & \cellcolor[HTML]{E1EAFF}\uline{0.1859}$^\ddagger$ & \cellcolor[HTML]{E1EAFF}\textbf{0.2296}$^\dagger$ & \cellcolor[HTML]{E1EAFF}\uline{0.1846}$^\dagger$ & \cellcolor[HTML]{E1EAFF}\uline{0.3480}$^\dagger$ & \cellcolor[HTML]{E1EAFF}\uline{0.1901} & \cellcolor[HTML]{E1EAFF}\uline{0.2609}$^\ddagger$ \\
\multirow{-4}{*}{\textit{Grok-4.1-fast}} & \cellcolor[HTML]{E1EAFF}\textbf{\agentname (Ours)} & \cellcolor[HTML]{E1EAFF}\ding{51} & \cellcolor[HTML]{E1EAFF}\textbf{0.4734}$^\ddagger$ & \cellcolor[HTML]{E1EAFF}\textbf{0.2321}$^\ddagger$ & \cellcolor[HTML]{E1EAFF}\uline{0.2236}$^\dagger$ & \cellcolor[HTML]{E1EAFF}\textbf{0.1861}$^\dagger$ & \cellcolor[HTML]{E1EAFF}0.3397 & \cellcolor[HTML]{E1EAFF}\textbf{0.2729}$^\ddagger$ & \cellcolor[HTML]{E1EAFF}\textbf{0.2880}$^\ddagger$ \\
\bottomrule
\end{tabular}%
}
\caption{Experiments results on MIMI-IV-Common subset. All the performances are measured by the F1 score. `*' indicates the model uses 4-bit AWQ quantization. The best results are \textbf{Bold}, while the second best results are \uline{underlined}. $\dagger$ and $\ddagger$ indicate the p-value $< 0.05$ and $< 0.01$ comparing with the ReSum method, respectively.}
\vspace{2 pt}
\label{tab: main results}
\end{table*}

\section{Experiments}
\subsection{Setting}
All the details of the experimental setting, including baseline models, agent methods, and experimental details, can be found in Appendix~\ref{appendix: experimental setting}.

\subsection{Main Results}
Table~\ref{tab: main results} demonstrates the consistent superiority of \agentname across all settings. Even without the evolving mechanism, our method outperforms competitive baselines. A key observation is the instability of the ReSum baseline on stronger backbones. While ReSum aids smaller models like Qwen3-Next-80B, it significantly underperforms standard ReAct on highly capable models such as Grok-4.1-fast (0.2237 vs 0.2501). This indicates that the critical information loss inherent in unidirectional compression outweighs its benefits for strong reasoners. In contrast, \agentname overcomes this limitation by retaining access to the full history, ensuring robust performance improvements regardless of the backbone's capability.

The results also highlight the general instability of evolving strategies within the \benchname environment. Reasoning Bank fails to improve weaker models and only yields benefits on larger parameters like the 80B model, while ReflectTool shows negligible gains because its atomic tool-specific experience is ineffective for complex context synthesis. Conversely, the evolving variant of \agentname achieves the highest average score of 0.2880. This success suggests that the high-quality context compression provided by the retrospective mechanism amplifies the utility of retrieved experiences, allowing the agent to leverage historical insights more effectively than other approaches.

\subsection{Cross-subset Validation}
To investigate the generalization capability of \agentname, we conducted cross-subset validation on MIMIC-IV-Rare and MIMIC-III. As detailed in Table~\ref{tab: cross data results}, \agentname exhibits superior robustness against both types of distribution shifts. In the MIMIC-IV-Rare subset, our method establishes a clear lead over ReSum and ReAct, effectively handling low-prevalence diagnoses where standard patterns often fail. The \textit{MIMIC-III} benchmark reveals a sharper contrast in adaptability. While baselines like ReflectTool demonstrate improved resilience, ReSum suffers a significant performance degradation. This decline suggests that ReSum's unidirectional summaries may be brittle to systemic format changes. Conversely, \agentname successfully mitigates this issue to maintain state-of-the-art performance, demonstrating that the retrospective mechanism captures generalizable medical logic rather than merely superior in a specific database.

\subsection{Ablation Experiments}
To validate the individual contributions of our framework's core components, we conducted comprehensive ablation studies on the Qwen3-30B-A3B backbone, as detailed in Table~\ref{tab: ablation}. Introducing the retrospective mechanism significantly enhances performance compared with Resum; applying it specifically to the Actor or Summarizer yields scores of 0.1876 and 0.1798, respectively, while the complete base \agentname (blue row) combining both achieves a synergistic gain to 0.1957. Building upon this, incorporating evolving optimization provides substantial benefits with the highest peak performance of 0.2117. A deeper analysis of the retrospective mechanism is further shown in Section~\ref{sec: effectiveness of retro}.

\begin{table*}[!t]
\centering
\resizebox{\textwidth}{!}{%
\begin{tabular}{llcccccccc}
\toprule
\textbf{EHR Database} & \textbf{Methods} & \textbf{Evolved} & \textbf{Diagnoses} & \textbf{Labevents} & \textbf{Microbiology} & \textbf{Prescriptions} & \textbf{Procedures} & \textbf{Transfers} & \textbf{Avg.} \\
\midrule
 & ReAct & \ding{55} & 0.0863 & \uline{0.0757} & 0.0550 & 0.0516 & 0.2645 & 0.2196 & 0.1255 \\
 & Reflexion & \ding{55} & 0.1248 & 0.0607 & 0.0447 & 0.0574 & 0.2492 & 0.1848 & 0.1203 \\
 & ReSum & \ding{55} & 0.1807 & 0.0609 & \uline{0.0579} & 0.0857 & 0.2812 & \uline{0.2282} & 0.1491 \\
 & ReflecTool (CS) & \ding{51} & 0.1653 & 0.0444 & \textbf{0.0597} & 0.0465 & 0.2506 & 0.2137 & 0.1300 \\
 & ReflecTool (IR) & \ding{51} & 0.0947 & 0.0446 & 0.0551 & 0.0361 & \uline{0.2826} & 0.2123 & 0.1209 \\
 & ReasoningBank & \ding{51} & 0.0309 & \textbf{0.0783} & 0.0540 & 0.0349 & 0.2570 & 0.1859 & 0.1068 \\
 & \cellcolor[HTML]{E1EAFF}\textbf{\agentname (Ours)} & \cellcolor[HTML]{E1EAFF}\ding{55} & \cellcolor[HTML]{E1EAFF}\uline{0.1950}$^\dagger$ & \cellcolor[HTML]{E1EAFF}0.0530 & \cellcolor[HTML]{E1EAFF}0.0519 & \cellcolor[HTML]{E1EAFF}\uline{0.0921} & \cellcolor[HTML]{E1EAFF}0.2719 & \cellcolor[HTML]{E1EAFF}\textbf{0.2400}$^\ddagger$ & \cellcolor[HTML]{E1EAFF}\uline{0.1506} \\
\multirow{-8}{*}{\textit{MIMIC-IV-Rare}} & \cellcolor[HTML]{E1EAFF}\textbf{\agentname (Ours)} & \cellcolor[HTML]{E1EAFF}\ding{51} & \cellcolor[HTML]{E1EAFF}\textbf{0.2325}$^\ddagger$ & \cellcolor[HTML]{E1EAFF}0.0700$^\dagger$ & \cellcolor[HTML]{E1EAFF}0.0536 & \cellcolor[HTML]{E1EAFF}\textbf{0.0933}$^\dagger$ & \cellcolor[HTML]{E1EAFF}\textbf{0.2991}$^\dagger$ & \cellcolor[HTML]{E1EAFF}0.2194 & \cellcolor[HTML]{E1EAFF}\textbf{0.1613}$^\dagger$ \\
\midrule
& ReAct & \ding{55} & {\color[HTML]{1F2329} 0.0395} & {\color[HTML]{1F2329} 0.1418} & {\color[HTML]{1F2329} \textbf{0.1871}} & {\color[HTML]{1F2329} 0.0841} & {\color[HTML]{1F2329} 0.0116} & 0.3272 & {\color[HTML]{1F2329} 0.1319} \\
 & Reflexion & \ding{55} & 0.0417 & 0.1340 & 0.1710 & 0.0715 & 0.0133 & 0.2656 & 0.1162 \\
 & ReSum & \ding{55} & 0.0502 & 0.1417 & \uline{0.1764} & \uline{0.0851} & 0.0185 & 0.2485 & 0.1201 \\
 & ReflecTool (CS) & \ding{51} & 0.0586 & 0.1956 & 0.1595 & 0.0782 & 0.0298 & \textbf{0.3689} & 0.1484 \\
 & ReflecTool (IR) & \ding{51} & 0.0468 & 0.1875 & 0.1602 & 0.0780 & 0.0282 & 0.3576 & 0.1431 \\
 & ReasoningBank & \ding{51} & 0.0602 & \textbf{0.2087} & 0.1736 & 0.0815 & 0.0170 & 0.1722 & 0.1189 \\
 & \cellcolor[HTML]{E1EAFF}\textbf{RetroSum (Ours)} & \cellcolor[HTML]{E1EAFF}\ding{55} & \cellcolor[HTML]{E1EAFF}\uline{0.0705}$^\ddagger$ & \cellcolor[HTML]{E1EAFF}\uline{0.1970}$^\ddagger$ & \cellcolor[HTML]{E1EAFF}0.1510$^\dagger$ & \cellcolor[HTML]{E1EAFF}\textbf{0.0891} & \cellcolor[HTML]{E1EAFF}\uline{0.0337}$^\ddagger$ & \cellcolor[HTML]{E1EAFF}\uline{0.3627}$^\ddagger$ & \cellcolor[HTML]{E1EAFF}\uline{0.1507}$^\ddagger$ \\
\multirow{-8}{*}{\textit{MIMIC-III}} & \cellcolor[HTML]{E1EAFF}\textbf{RetroSum (Ours)} & \cellcolor[HTML]{E1EAFF}\ding{51} & \cellcolor[HTML]{E1EAFF}\textbf{0.0869}$^\ddagger$ & \cellcolor[HTML]{E1EAFF}0.1860$^\ddagger$ & \cellcolor[HTML]{E1EAFF}0.1691 & \cellcolor[HTML]{E1EAFF}0.0807 & \cellcolor[HTML]{E1EAFF}\textbf{0.0473}$^\ddagger$ & \cellcolor[HTML]{E1EAFF}0.3570$^\ddagger$ & \cellcolor[HTML]{E1EAFF}\textbf{0.1545}$^\ddagger$ \\
\bottomrule
\end{tabular}%
}
\caption{Results of Qwen3-30B-A3B on OOD subsets. The best results are \textbf{Bold}, while the second best results are \uline{underlined}. $\dagger$ and $\ddagger$ indicate the p-value $< 0.05$ and $< 0.01$ comparing with the ReSum method, respectively.}
\label{tab: cross data results}
\end{table*}

\begin{table*}[!t]
\centering
\resizebox{\textwidth}{!}{%
\begin{tabular}{ccccccccccc}
\toprule
\multicolumn{2}{c}{\textbf{Actor}} & \multicolumn{2}{c}{\textbf{Summarizer}} &  &  &  &  &  &  &  \\
\textbf{Retrospect} & \textbf{Evolved} & \textbf{Retrospect} & \textbf{Evolved} & \multirow{-2}{*}{\textbf{Diagnoses}} & \multirow{-2}{*}{\textbf{Labevents}} & \multirow{-2}{*}{\textbf{Microbiology}} & \multirow{-2}{*}{\textbf{Prescriptions}} & \multirow{-2}{*}{\textbf{Procedures}} & \multirow{-2}{*}{\textbf{Transfers}} & \multirow{-2}{*}{\textbf{Avg.}} \\
\midrule
\rowcolor[HTML]{DEE0E3}
\ding{55} & \ding{55} & \ding{55} & \ding{55} & 0.1753 & 0.0881 & 0.1166 & 0.1000 & 0.2717 & 0.2320 & 0.1639 \\
\ding{51} & \ding{55} & \ding{55} & \ding{55} & 0.2257$^\ddagger$ & 0.1041$^\dagger$ & 0.1418$^\ddagger$ & 0.1030 & 0.2932$^\dagger$ & 0.2577 & 0.1876$^\ddagger$ \\
\ding{55} & \ding{55} & \ding{51} & \ding{55} & 0.1759 & 0.1097$^\dagger$ & \textbf{0.1710}$^\ddagger$ & 0.1119$^\dagger$ & 0.2320$^\ddagger$ & 0.2782$^\dagger$ & 0.1798$^\ddagger$ \\
\rowcolor[HTML]{E1EAFF} 
\ding{51} & \ding{55} & \ding{51} & \ding{55} & 0.2368$^\ddagger$ & 0.0977$^\ddagger$ & \uline{0.1584}$^\ddagger$ & 0.1039 & 0.2983$^\dagger$ & 0.2791 & 0.1957$^\ddagger$ \\
\ding{51} & \ding{51} & \ding{51} & \ding{55} & \uline{0.2505}$^\ddagger$ & \textbf{0.1387}$^\ddagger$ & 0.1541$^\ddagger$ & \textbf{0.1354}$^\dagger$ & 0.2853 & \uline{0.2982}$^\ddagger$ & \uline{0.2104}$^\ddagger$ \\
\ding{51} & \ding{55} & \ding{51} & \ding{51} & 0.2437$^\ddagger$ & \uline{0.1151}$^\ddagger$ & 0.1565$^\ddagger$ & 0.1082 & \textbf{0.3234}$^\ddagger$ & 0.2919$^\ddagger$ & 0.2065$^\ddagger$  \\
\rowcolor[HTML]{E1EAFF} 
\ding{51} & \ding{51} & \ding{51} & \ding{51} & \textbf{0.2514}$^\ddagger$ & 0.1092$^\dagger$ & 0.1568$^\ddagger$ & \uline{0.1289}$^\dagger$ & \uline{0.3068}$^\dagger$ & \textbf{0.3171}$^\ddagger$ & \textbf{0.2117}$^\ddagger$ \\
\bottomrule
\end{tabular}%
}
\caption{Ablation results of Qwen3-30B-A3B on MIMIC-IV-Common subset. The row in \colorbox[HTML]{DEE0E3}{gray} and \colorbox[HTML]{E1EAFF}{blue} indicates the results of ReSum and \agentname, respectively. The best results are \textbf{Bold}, while the second best results are \uline{underlined}. $\dagger$ and $\ddagger$ indicate the p-value $< 0.05$ and $< 0.01$ comparing with the ReSum method, respectively.}
\label{tab: ablation}
\end{table*}

\section{Analysis}
In this section, all the experiments are conducted on the diagnoses task in MIMIC-IV-Common. More experimental results, like tool analysis and caes study, can be found in Appendix~\ref{appendix: experiments}.

\subsection{Effectiveness of Retrospective Mechanism}
\label{sec: effectiveness of retro}
\begin{figure}[!t]
    \centering
        \includegraphics[width=0.6\linewidth]{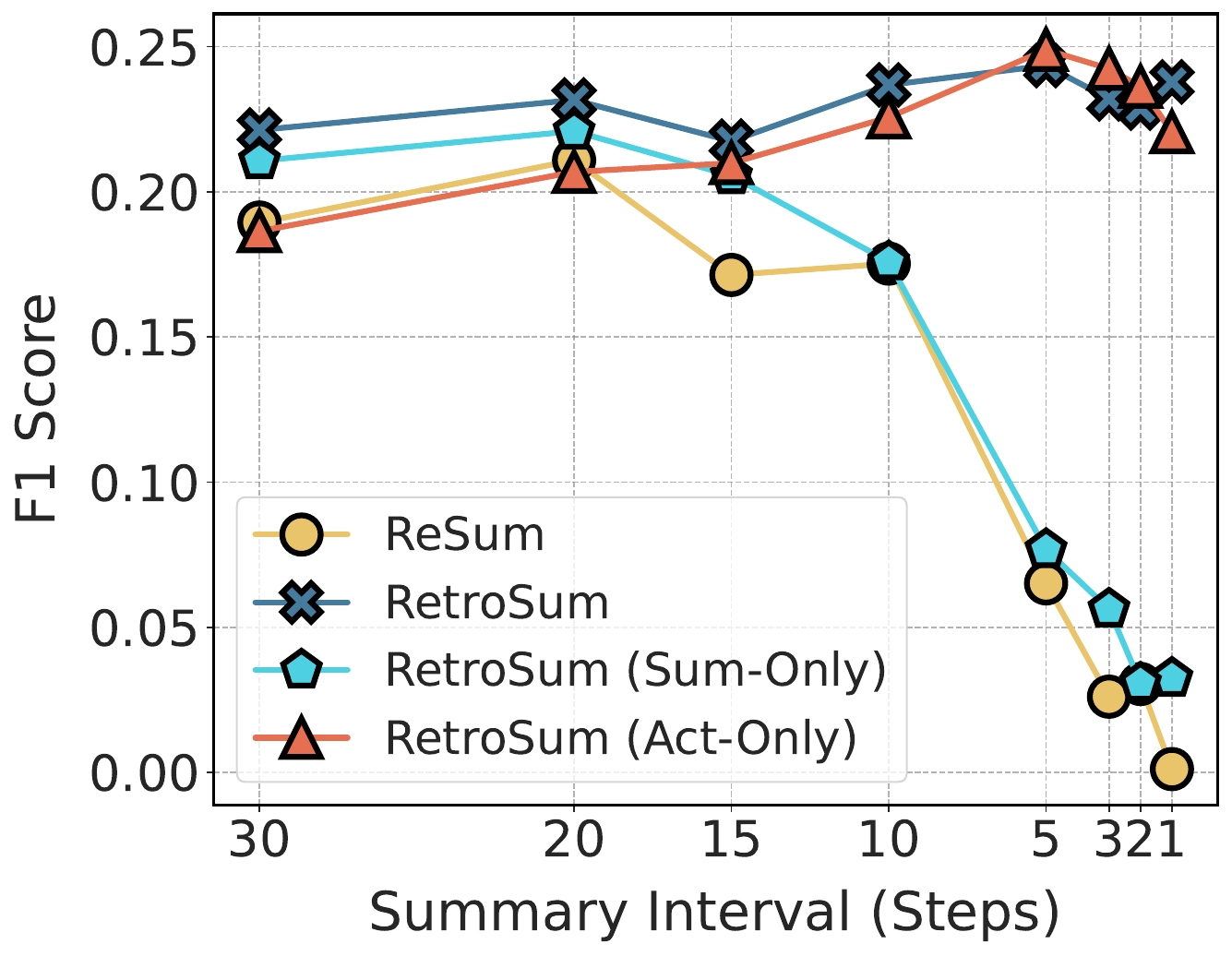}
        \caption{Impact of summarization interval on agent performance. The retrospective mechanism is applied solely to the Summarizer (Sum-Only) or the Actor (Act-Only) across varying frequencies.} 
        \label{fig: summary_interval}
\end{figure}

To dissect the contributions of the retrospective mechanism, we evaluated its independent application to the Summarizer (Sum-Only) and Actor (Act-Only) across intervals ranging from 30 down to 1 step, as shown in Figure~\ref{fig: summary_interval}. The results reveal distinct dominant roles depending on summarization frequency. At small intervals (e.g., $w \le 5$), Act-Only is crucial and drives performance by maintaining immediate reasoning coherence against frequent context interruptions. Conversely, at larger intervals (e.g., $w \ge 15$), Sum-Only becomes dominant, ensuring distant critical information is retained over long horizons. By synergizing these complementary strengths, the complete \agentname framework secures robust performance that consistently outperforms the ReSum baseline across the entire spectrum of summarization frequencies.

\begin{figure}[!t]
    \centering
    \includegraphics[width=\linewidth]{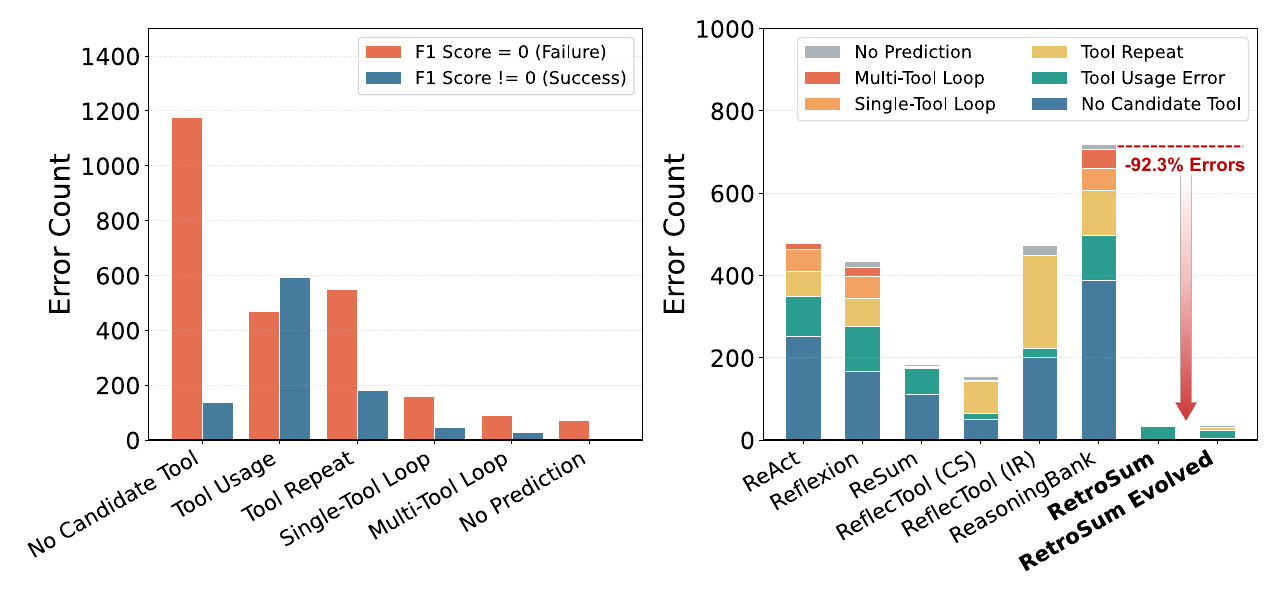}
    \caption{Error statistics on the diagnoses task. The left panel displays the distribution of specific error types across unsuccessful and successful trajectories. The right panel compares the count of errors committed by different agent frameworks.}
    \label{fig: error_analysis}
\end{figure}

\begin{figure*}[t] 
    \centering
    
    \begin{minipage}[t]{0.32\textwidth}
        \centering
        \includegraphics[width=\linewidth]{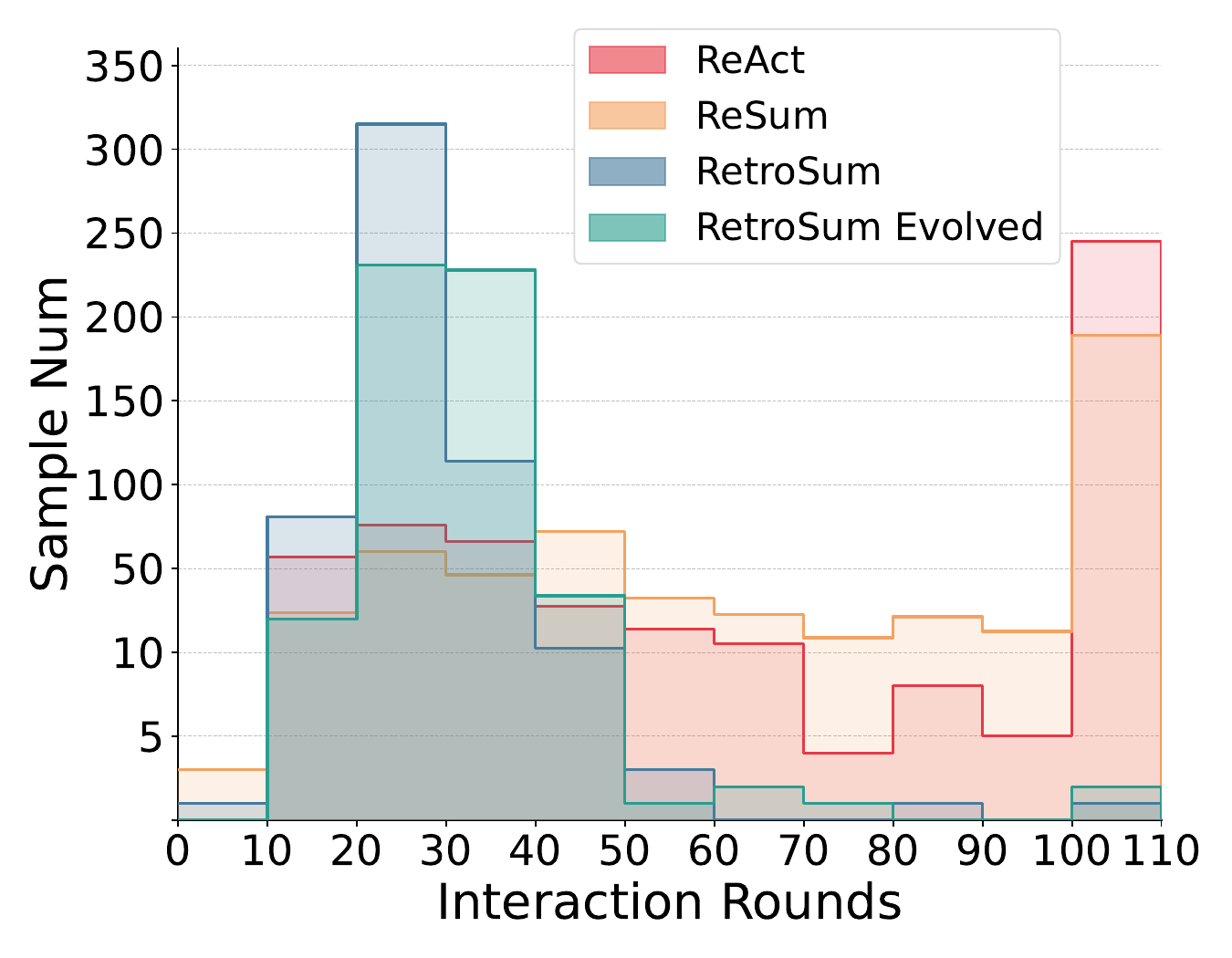}
        \caption{Distribution of interaction turns across different agent methods.} 
        \label{fig: turn_analysis}
        \label{fig: turn}
    \end{minipage}
    \hfill 
    \begin{minipage}[t]{0.32\textwidth}
        \centering
        \includegraphics[width=.96\linewidth]{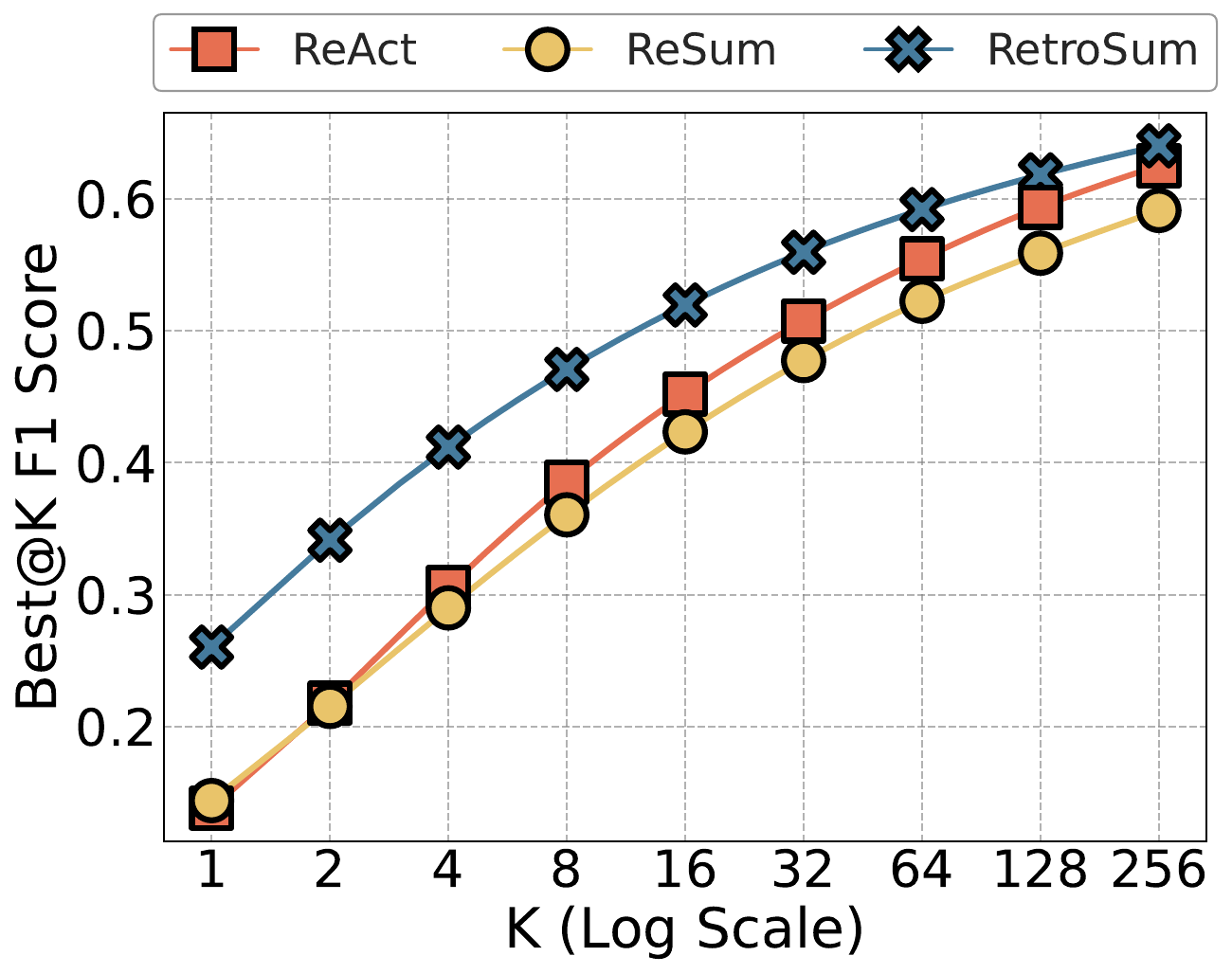}
        \caption{Test-time scaling performance evaluated using Best@K F1 Score.}
        \label{fig: scaling}
    \end{minipage}
    \hfill 
    \begin{minipage}[t]{0.32\textwidth}
        \centering
        \includegraphics[width=\linewidth]{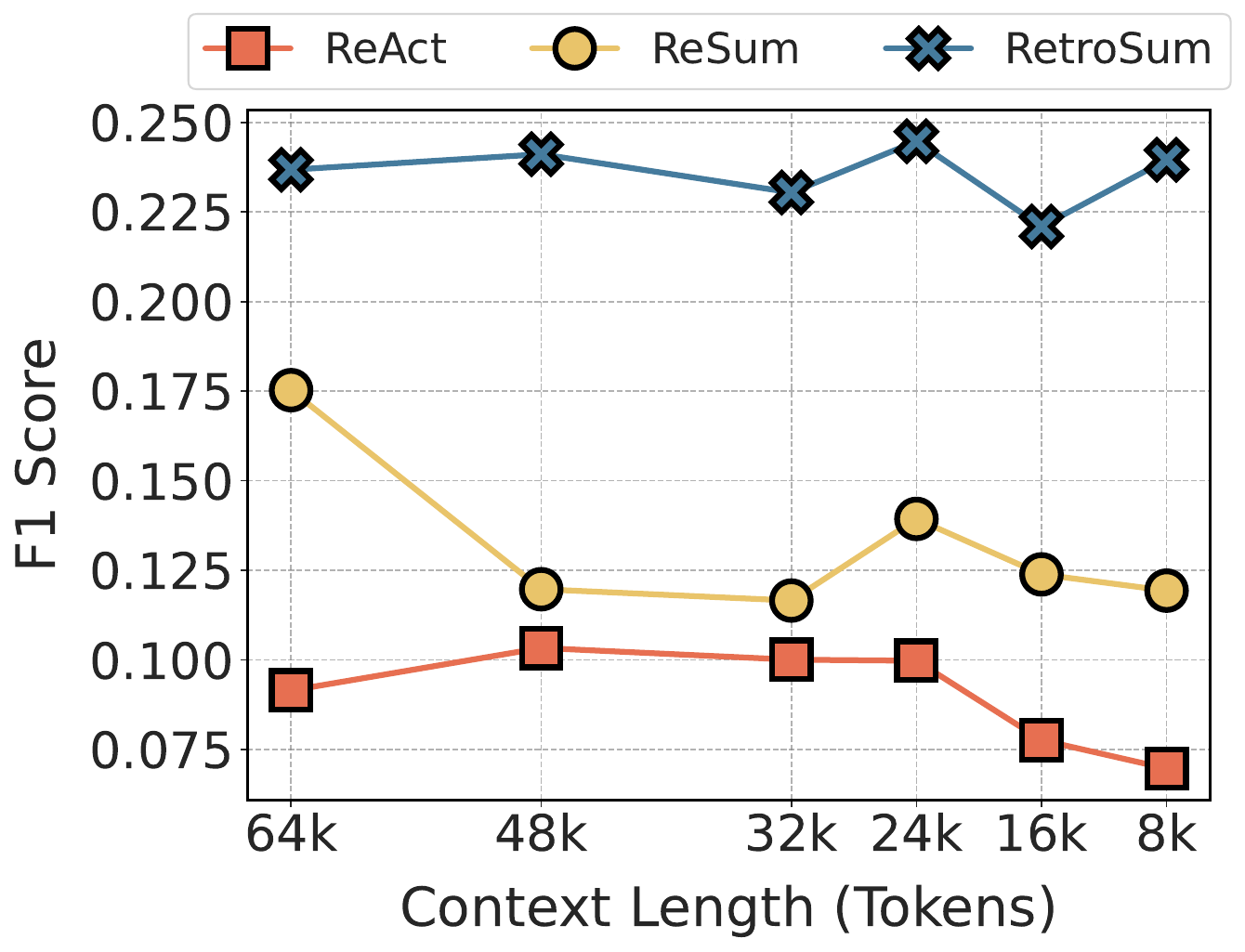}
        \caption{Performance sensitivity to maximum context length limited from 64k to 8k tokens.}
        \label{fig: context_length}
    \end{minipage}
\end{figure*}

\subsection{Error Analysis}
To deepen our understanding, we categorized failures into six types (detailed in Appendix~\ref{appendix: error_definition}). As shown in Figure \ref{fig: error_analysis}, failed trajectories are predominantly plagued by `No Candidate Tool' errors and repetitive behaviors, indicating that EHR noise frequently disrupts reasoning context. While ReSum reduces repetitive errors compared to ReAct, it remains prone to tool-related failures due to information loss. In contrast, \agentname achieves a substantial reduction in total errors across all categories. Notably, the \textbf{Evolved \agentname} variant further enhances tool utilization, resulting in a 92.3\% reduction in total errors compared to ReasoningBank. These findings validate that our retrospective mechanism effectively mitigates critical reasoning failures caused by complex, long-horizon EHR interactions.

\subsection{Efficiency Analysis}
\label{subsec: Turn Analysis}



To evaluate reasoning efficiency, we analyzed the distribution of interaction turns in Figure \ref{fig: turn_analysis}~(further analysis in Appendix~\ref{appendix: turn analysis}). The results highlight a critical bottleneck in baselines: ReSum frequently exhausts the maximum 100-turn limit, indicating that its lossy unidirectional summarization traps agents in redundant information-seeking loops to recover missing context. In contrast, \agentname shifts the distribution peak to the efficient 20-40 turn range. By utilizing the retrospective mechanism to retain salient history, our method effectively eliminates navigational dead-ends. Crucially, this reasoning efficiency translates to resource economy. Although the retrospective process introduces specific overhead per step, the drastic reduction in total interaction turns effectively outweighs this cost. Consequently, \agentname achieves superior performance while requiring significantly fewer tokens and less inference time than baselines (as detailed in Appendix~\ref{appendix: computational consumption analysis}).

\subsection{Test-time Scaling}

To explore the upper bounds of agent capability, we investigate the impact of test-time scaling using the Best@K F1 Score (see Appendix~\ref{appendix: metrics} for formal definition). As illustrated in Figure \ref{fig: scaling}, \agentname consistently outperforms both ReAct and ReSum baselines across the entire spectrum of $K$. The superior performance at Best@1 (equivalent to the average F1 score across 256 samples) attests to the robustness of our retrospective mechanism in consistently generating high-quality trajectories compared to baselines. Furthermore, the substantial performance gains observed with increased sampling budgets indicate that current models possess significant potential capacity to solve these complex clinical problems. This confirms that the \benchname task, while challenging, is fundamentally solvable and its design is reasonable.

\subsection{Impact of Context Length}


To evaluate the robustness of our framework under memory constraints, we varied the maximum context length from 64k down to 8k tokens, as shown in Figure \ref{fig: context_length}. A clear downward trend is observed for baseline methods; both ReAct and ReSum suffer significant performance degradation as the context window tightens, indicating that standard sliding windows and unidirectional summaries fail to retain critical information when space is limited. Conversely, \agentname exhibits remarkable stability, maintaining high F1 scores even at the most restrictive 8k limit. This robustness stems from the retrospective mechanism, which enables the Actor to preserve complete reasoning logic regardless of window size. By periodically re-evaluating the full interaction history to distill essential cross-temporal dependencies into the summary, \agentname ensures the reasoning chain remains unbroken and information-dense, effectively neutralizing the negative impact of restricted context windows.

\section{Conclusions}
In this work, we bridge the gap between idealized experimental settings and realistic clinical environments by presenting \benchname, a benchmark necessitating complex decision-making within raw, high-noise databases. Based on this, we propose \agentname, which adopt the retrospective mechanism with an evolving experience strategy to capture latent cross-temporal correlations. Empirical validation shows retrospective mechanisms are essential for unlocking the potential of clinical agents in EHR-based reasoning decision-making tasks.

\section*{Limitations}
Despite the robust performance of \agentname established in this work, several limitations remain. First, our evaluation relies primarily on the MIMIC-IV and MIMIC-III datasets. While these databases serve as the gold standard for critical care research, they are inherently sourced from a single medical center, which may not fully capture the diverse administrative protocols or demographic variations found in broader global healthcare systems. Second, our current framework is designed specifically for textual clinical notes and structured tabular data. It does not yet possess the capability to directly analyze pixel-level medical imaging (e.g., raw CT scans) or high-frequency physiological waveforms, relying instead on textual reports. Future work will focus on extending the agent's capabilities to multimodal data synthesis and exploring multi-center generalization.



\bibliography{custom}

\newpage
\appendix
\section{Related Works}
Recent advancements in Large Language Models (LLMs) for Electronic Health Records (EHRs) have progressed from foundational instruction tuning to the development of autonomous agents and specialized benchmarks. Early efforts focused on aligning general-purpose models with the longitudinal and heterogeneous nature of clinical data; for instance, MIMIC-Instr~\cite{wu2024instruction} and MEDALIGN~\cite{fleming2024medalign} introduced large-scale instruction datasets to bridge the gap between raw clinical logs and natural language understanding. These foundations were subsequently enhanced by frameworks addressing specific reasoning dimensions, such as TIMER~\cite{cui2025timer}, which targets temporal dependencies in patient histories, and EHR-R1~\cite{liao2025ehr}, which synthesizes reasoning chains to improve complex decision-making. Concurrently, the field has shifted towards agentic systems capable of active tool utilization. EHRAgent~\cite{shi-etal-2024-ehragent} and ReflecTool~\cite{DBLP:conf/acl/LiaoJW025} demonstrated that equipping LLMs with code execution interfaces and reflective memory significantly outperforms standard prompting on tabular reasoning tasks. This transition to agency extends to real-world integration, where EHR-MCP~\cite{masayoshi2025ehr} and EHRMIND~\cite{DBLP:journals/corr/abs-2505-09388} explore standard-compliant deployment and verifiable reinforcement learning to ensure reliability in live clinical settings. To rigorously evaluate these interactive capabilities, novel benchmarks like MedAgentBench~\cite{jiang2025medagentbench} and FHIR-AgentBench~\cite{lee2025fhir} have been established, moving beyond static question-answering to assess agent planning, multi-step tool usage, and interoperability within realistic virtual EHR environments.

\section{Data Curation}
\label{appendix: data_curation}

This section outlines the data curation methodology used to convert raw Electronic Health Records (EHR) into a structured format. The process involves selecting data sources, with MIMIC-IV serving as the primary corpus and MIMIC-III used for cross-subset validation. We constructed patient-level time series, enriched medical codes with semantic mappings, standardized storage formats, and developed a unified toolbox to utilize the curated data for executing clinical tasks. The following subsections describe the processing steps ensuring data integrity and chronological accuracy.

\subsection{Data Preprocessing and Formatting}
\label{subsec:preprocessing}

We reformatted the raw MIMIC-IV and MIMIC-III dataset to construct accurate patient-level time series. This involved extracting all events for each patient and sorting them by timestamp with second-level precision.

\paragraph{Temporal Alignment and Imputation.}
The raw dataset lacks second-level timestamps for tables such as \texttt{diagnoses\_icd}, \texttt{procedures\_icd}, and diagnosis events in the Emergency Department (ED). To enable sequential modeling, we imputed these timestamps by linking events to their corresponding admissions. For both standard \texttt{diagnoses\_icd} and ED diagnoses, we set the timestamp to one minute prior to the discharge time of the associated admission or ED stay. For \texttt{procedures\_icd} events that only provide day-level resolution, we assigned a default timestamp of 23:59:59 on the recorded day to maintain logical ordering.

\paragraph{Semantic Enrichment.}
We mapped high-dimensional medical codes to broader clinical categories to improve representation. International Classification of Diseases~(ICD) codes in diagnosis and procedure tables were mapped to Clinical Classification Software (CCS) categories using the `ICD-to-CCS' script\footnote{From ICD-9 to CCS: \url{https://hcup-us.ahrq.gov/toolssoftware/ccs/ccs.jsp}.}\footnote{From ICD-10 to CCSR: \url{https://hcup-us.ahrq.gov/toolssoftware/ccsr/dxccsr.jsp}.}. Similarly, National Drug Codes (NDC) in prescription tables were converted to Anatomical Therapeutic Chemical (ATC) codes adopting the off-the-shelf mapping script\footnote{From NDC to ATC: \url{https://github.com/sunlabuiuc/PyHealth}.}. These mappings consolidate sparse code representations into hierarchically structured features.

\paragraph{Textual Integration and Leakage Mitigation.}
We standardized textual data availability by moving information explicitly available at admission, such as Social History and Chief Complaints, from the discharge summary to the admission event. To maintain dataset consistency, we restricted our cohort to patients possessing valid text records in their discharge summaries. We also enforced strict leakage control by removing columns containing future administrative or outcome information. In the \texttt{admissions} table, we excluded \texttt{dischtime}, \texttt{deathtime}, \texttt{discharge\_location}, \texttt{edouttime}, and \texttt{hospital\_expire\_flag}. Similarly, we removed \texttt{last\_careunit}, \texttt{outtime}, and \texttt{los} from \texttt{icustays}, as well as \texttt{outtime} and \texttt{disposition} from \texttt{edstays}. Finally, we masked pharmacy artifacts in prescription events that implicitly reveal future details.

\paragraph{Label Space Construction.}
To facilitate effective clinical reasoning across both MIMIC-IV and MIMIC-III, we addressed the challenge of excessive granularity in the raw label space. We employed three distinct strategies based on data characteristics. First, for \texttt{diagnoses} and \texttt{procedures}, we adopted the Clinical Classification Software (CCS) taxonomy to aggregate granular codes into clinically coherent categories. This transformation yielded a significant reduction, exemplified by MIMIC-IV. The label space was compressed from 109,775 raw diagnosis codes to 283 CCS categories, and from 85,257 procedure codes to 231 categories. Similarly, for \texttt{prescriptions} in MIMIC-IV, we projected National Drug Codes (NDC) onto the Anatomical Therapeutic Chemical (ATC) system, condensing 1,086,608 raw items into 1,813 pharmacological groups.

For domains lacking standardized mappings, we utilized statistical aggregation or metadata definitions. Specifically, due to the absence of consistent NDC records in MIMIC-III \texttt{prescriptions}, as well as for \texttt{microbiologyevents} and \texttt{transfers} in both datasets, label spaces were constructed by aggregating all unique items observed in the records. For \texttt{labevents}, the candidate list was derived directly from the standardized definition table (\texttt{d\_labitems}), ensuring coverage of all valid laboratory tests.

\paragraph{Storage Structure.}
Processed data is stored in a patient-centric format where the complete longitudinal history of each patient is contained within a single SQLite database file. This structure supports modular access and efficient data retrieval.

\subsection{Auxiliary Resources}
\label{subsec:auxiliary}

We constructed Reference and Candidates databases to facilitate data interpretation and define the output space for predictive tasks.

\paragraph{Reference Database.}
The Reference database functions as a dictionary that maps medical codes to item names or descriptions using metadata from the \texttt{diagnoses}, \texttt{procedures}, \texttt{prescriptions}, and \texttt{labitems} tables in MIMIC-IV. This resource allows the retrieval of specific item names associated with patient EHR records.

\paragraph{Candidates Database.}
The Candidates database defines the answer space for the six evaluation tasks. Candidate lists for \texttt{diagnoses}, \texttt{procedures}, \texttt{prescriptions}, and \texttt{labitems} were extracted directly from the Reference database. For domains without dedicated metadata tables, specifically \texttt{microbiologyevents} and \texttt{transfers}, we aggregated all unique items observed across the MIMIC-IV and MIMIC-III datasets to form the final candidate sets. This process resulted in six candidate tables covering the required label space. We also provide schema description files and a database link reference to support system integration.

\subsection{Sample Construction}
\label{subsec:cohort_selection}

To construct a robust benchmark from the massive MIMIC-IV database ($\sim$ 300k patients) and the smaller MIMIC-III database($\sim$ 1k patients), we employ a stratified sampling strategy designed to balance label diversity and task difficulty.

\paragraph{MIMIC-IV Construction.}
For each clinical task, we first identify valid event occurrences for each patient. We define a sample instance by setting the prediction timestamp $t$ to one minute prior to the event time and identifying the ground-truth label set $Y^*$. Subsequently, we temporally censor the patient's history by removing any clinical observations recorded after the prediction time $t$. To ensure the benchmark covers both prevalent conditions and long-tail rare cases, we implement a Label-wise Weighted Sampling strategy. We calculate a sampling weight $w_S$ for each candidate sample $S$, inversely proportional to the frequency of its constituent labels. Formally, for a sample $S$ with a ground-truth label set $Y^* = \{y_1, y_2, \dots, y_N\}$, the weight is defined as:
\begin{equation}
w_S = \frac{1}{|Y^*|} \sum_{y \in Y^*} \frac{1}{\text{Count}(y)}
\end{equation}
where $\text{Count}(y)$ represents the global frequency of label entity $y$ within the specific task type.We then compute the mean weight $\bar{w}$ across all candidates to stratify the dataset. Samples with $w_S \le \bar{w}$ are categorized into the \textit{Common} pool (high-frequency labels), while those with $w_S > \bar{w}$ are assigned to the \textit{Rare} pool (low-frequency labels). Within each pool, we perform weighted random sampling using $w_S$ as the probability distribution. This ensures that even within the \textit{Common} or \textit{Rare} subsets, the selected samples are uniformly distributed across the label space, maximizing diversity. Finally, we sample 600 samples for each type of task and filter out samples lacking admission records to ensure data completeness, resulting in the final \textit{MIMIC-IV-Common} and \textit{MIMIC-IV-Rare} benchmarks.

\paragraph{MIMIC-III Construction.}Given the significantly smaller scale of MIMIC-III, we adopt a direct sampling approach. We randomly sample instances from the valid patient pool. However, to account for the systemic complexity differences, we selectively increase the sample density for high-difficulty tasks, specifically \textit{Diagnoses} and \textit{Procedures}, to ensure the evaluation metric remains statistically significant for these challenging reasoning scenarios.

\subsection{Dataset Statistics}
\label{subsec:statistics}

Table \ref{tab: Data Statistics} presents a comprehensive statistical summary of the curated benchmarks. The datasets are categorized into \textit{MIMIC-IV Common}, \textit{MIMIC-IV Rare}, and \textit{MIMIC-III} subsets to evaluate model performance across different prevalence distributions and data sources. 

A defining characteristic of the \textit{MIMIC-IV} subsets is the high complexity of the input data. On average, the model must process longitudinal patient histories spanning over a year, comprising thousands of temporal records. This necessitates robust capabilities in modeling long-term dependencies within heterogeneous EHR tables.

In distinct contrast, the \textit{MIMIC-III} subset exhibits a fundamentally different structural challenge. Unlike MIMIC-IV, MIMIC-III covers a significantly shorter time span (approximately 1/20) yet retains a disproportionately large volume of records (roughly 1/2). This discrepancy indicates a substantially higher recording density, which inevitably introduces greater redundancy and noise. Consequently, as evidenced by the comparative results in Table \ref{tab: cross data results}, MIMIC-III presents a more formidable challenge than MIMIC-IV. We attribute this difficulty not to data scarcity, but to the heightened complexity of information retrieval: the model is required to filter through highly dense and noisy sequences to extract relevant clinical signals effectively.

\begin{table*}[!t]
\centering
\resizebox{\textwidth}{!}{%
    \begin{tabular}{llcccccc}
        \toprule
        \textbf{Dataset} & \textbf{Task} & \textbf{\# Cases} & \textbf{\# Cand.} & \textbf{Avg. Label} & \textbf{Avg. Tables} & \textbf{Avg. Recs} & \textbf{Avg. Days} \\
        \midrule
        
        \multirow{6}{*}{\textit{MIMIC-IV Common}}
          & Diagnoses          & 528 & 283  & 12.03 & 24.17 & 3041.10 & 370.84 \\
          & Labevents          & 589 & 1170 & 30.26 & 17.60 & 849.92  & 347.79 \\
          & Microbiologyevents & 549 & 171  & 5.59  & 21.04 & 1736.62 & 510.68 \\
          & Prescriptions      & 515 & 1813 & 7.65  & 24.03 & 1786.62 & 414.26 \\
          & Procedures         & 529 & 231  & 2.46  & 24.20 & 1797.61 & 384.50 \\
          & Transfers          & 538 & 38   & 1.00  & 25.83 & 3237.72 & 387.00 \\
        \midrule
        
        \multirow{6}{*}{\textit{MIMIC-IV Rare}} 
          & Diagnoses          & 559 & 283  & 6.10  & 20.95 & 1341.71 & 301.63 \\
          & Labevents          & 600 & 1170 & 18.58 & 17.04 & 1001.75 & 438.99 \\
          & Microbiologyevents & 565 & 171  & 4.29  & 19.18 & 1401.23 & 524.76 \\
          & Prescriptions      & 527 & 1813 & 7.27  & 23.64 & 1455.21 & 349.78 \\
          & Procedures         & 493 & 231  & 2.28  & 23.36 & 1642.98 & 404.20 \\
          & Transfers          & 487 & 38   & 1.00  & 24.80 & 2691.69 & 441.25 \\
        \midrule
        
        \multirow{6}{*}{\textit{MIMIC-III}} 
          & Diagnoses          & 600 & 283  & 11.03 & 11.57 & 1032.14 & 8.81  \\
          & Labevents          & 500 & 587  & 18.92 & 11.65 & 679.61  & 26.19 \\
          & Microbiologyevents & 500 & 63   & 3.23  & 11.73 & 643.05  & 21.23 \\
          & Prescriptions      & 500 & 1235 & 15.55 & 11.62 & 564.59  & 40.62 \\
          & Procedures         & 692 & 231  & 3.64  & 11.81 & 1275.52 & 55.92 \\
          & Transfers          & 500 & 5    & 1.00  & 11.76 & 550.49  & 31.34 \\       
        \bottomrule
    \end{tabular}%
}
\caption{Benchmark statistics for each \textbf{Dataset} and \textbf{Task} cohort (\textbf{\# Cases}). The table details the number of candidate answers for the task (\textbf{\# Cand.}) and the average size of label set (\textbf{Avg. Label}). Input complexity is further characterized by the average number of source tables (\textbf{Avg. Tables}), record volume (\textbf{Avg. Recs}), and longitudinal span (\textbf{Avg. Days}).}
\label{tab: Data Statistics}
\end{table*}

\section{Toolbox Construction}
\label{appendix: toolbox}

To enable autonomous agents to effectively interact with the curated clinical environment, we developed a comprehensive toolbox serving as the interface between the agent’s reasoning core and the structured data. The toolbox is categorized into five distinct functional modules: \textbf{Record}, \textbf{Candidate}, \textbf{Table}, \textbf{Inner}, and \textbf{Retrieval}. A detailed specification of the constituent tools and their parameters is provided in Table \ref{tab: Toolbox Overview}.

\paragraph{Record Interaction Tools.}
These tools facilitate granular access to the patient's longitudinal history by acting as a direct query interface to the EHR database. To retrieve specific information, the agent must specify the target table name along with precise filtering criteria. The tool supports queries based on temporal constraints to isolate events within specific time windows, textual content matching to identify records containing particular medical terms, and numerical value filtering to extract measurements meeting specific thresholds. This mechanism allows the agent to dynamically gather evidence from the patient's history without needing to load the entire database into context.

\paragraph{Candidate Alignment Tools.}
This category is critical for grounding the agent's free-form reasoning into the fixed output space defined by the Candidate Database. To ensure robust mapping between generated hypotheses and valid label entries, the tool employs a hybrid matching strategy. It first attempts direct keyword search and fuzzy string matching to handle exact matches and minor morphological variations. For more complex cases involving synonymy or terminological differences, the tool utilizes semantic alignment powered by the biomedical-specific embedding model \textbf{BioLORD-2023}~\citep{remy-etal-2023-biolord}. This vector-based retrieval enables the agent to identify the correct candidate even when there is no lexical overlap between the generated query and the standardized code description.

\paragraph{Schema Inspection Tools.}
To effectively navigate the relational structure of the EHR, the agent utilizes Schema Inspection tools to acquire metadata awareness. These tools allow the agent to query the definitions of the database structure itself, revealing which tables are available in the current patient's file and detailing the specific columns and data types within each table. By understanding the underlying schema, the agent can formulate syntactically correct queries for the Record Interaction tools and interpret the retrieved data with the correct semantic context.

\paragraph{Cognitive Management Tools (Inner).}
Unlike external interaction tools, this category governs the agent's internal control flow and decision-making process. The \texttt{Think} tool enables the agent to generate intermediate reasoning traces and formulate multi-step plans without triggering external environment actions. Once the agent has gathered sufficient evidence and reached a conclusion, the \texttt{Finish} tool serves as the termination signal, allowing the agent to end the trajectory and output the final prediction. This separation of reasoning and termination ensures a structured and verifiable thought process.

\paragraph{External Knowledge Retrieval.}
To augment the specific clinical data contained within the EHR, this tool connects the agent to a broader external knowledge base. By accepting natural language queries, it searches for relevant medical literature, guidelines, or definitions that are not present in the patient's records. In implementation, we use the data sources provided in MedRAG~\cite{xiong2024benchmarking}. This provides the agent with the necessary background knowledge to interpret complex medical conditions or rare procedures, thereby supporting more informed decision-making.

\begin{table*}[!t]
\centering
\renewcommand{\arraystretch}{1.2}
\resizebox{\textwidth}{!}{%
\begin{tabular}{clp{9cm}p{8cm}}
\toprule
\textbf{Category} & \textbf{Tool Name} & \textbf{Description} & \textbf{Parameters} \\
\midrule

\multirow{4}{*}[-1em]{\textit{Inner}} 
 & \texttt{think} & Designed to synthesize information gathered from preceding operations and to articulate the necessary subsequent actions. & \texttt{response} \\
 & \texttt{finish} & The final step in the reasoning process. Used only when all necessary data has been retrieved and the clinical prediction is ready. & \texttt{response} \\
\midrule

\multirow{8}{*}[-3.5em]{\textit{Record}} 
 & \texttt{get\_records\_by\_time} & Finds records in a EHR Table that fall within a given time range. & \texttt{subject\_id}, \texttt{table\_name}, \texttt{start\_time}, \texttt{end\_time} \\
 & \texttt{get\_event\_counts\_by\_time} & Calculates the number of events in all EHR Tables that fall within a given time range. & \texttt{subject\_id}, \texttt{start\_time}, \texttt{end\_time} \\
 & \texttt{get\_latest\_records} & Finds the latest timestamp and returns all EHR Table records that share that same timestamp in EHR Table. & \texttt{subject\_id}, \texttt{table\_name} \\
 & \texttt{get\_records\_by\_keyword} & Searches for all text-based columns of the specific EHR Table containing a specific keyword. & \texttt{subject\_id}, \texttt{table\_name}, \texttt{keyword} \\
 & \texttt{get\_records\_by\_value} & Finds records in a EHR Table where a given column's value is exact match for the keyword. & \texttt{subject\_id}, \texttt{table\_name},\texttt{column\_name}, \texttt{value} \\
 & \texttt{run\_sql\_query} & Executes a standard SQL query against the patient's EHR Table. & \texttt{subject\_id}, \texttt{sql\_query} \\
 & \texttt{get\_unique\_values} & Retrieves all unique values from a specified categorical column in an EHR table. & \texttt{subject\_id}, \texttt{table\_name}, \texttt{column\_name} \\
\midrule

\multirow{3}{*}[-2em]{\textit{Candidate}} 
 & \texttt{get\_candidates\_by\_keyword} & Searches for all text-based columns of the specific Candidate Table containing a specific keyword. & \texttt{table\_name}, \texttt{keyword} \\
 & \texttt{get\_candidates\_by\_fuzzy\_matching} & Finds similar items in a Candidate Table based on fuzzy matching. & \texttt{table\_name}, \texttt{keywords} \\
 & \texttt{get\_candidates\_by\_semantic\_similarity} & Performs semantic search using BioLORD-2023 embeddings to find semantically similar unique entities. & \texttt{table\_name}, \texttt{query} \\
\midrule

\multirow{3}{*}[-1.5em]{\textit{Table}} 
 & \texttt{get\_column\_names} & Retrieves all column names for a specified table for understanding the data. & \texttt{subject\_id}, \texttt{table\_name} \\
 & \texttt{get\_table\_names} & Retrieves the names of all available tables in the database, categorized into EHR tables and candidates tables.  & \texttt{subject\_id} \\
 & \texttt{get\_table\_description} & Retrieve EHR table description and column information from the hospital database schema. & \texttt{table\_name} \\
\midrule

\multirow{5}{*}[-1.5em]{\textit{Knowledge}} 
 & \texttt{retrieve\_pubmed} & Retrieve abstract of relevant biomedical documents from PubMed corpus given a query. & \texttt{query} \\
 & \texttt{retrieve\_textbooks} & Retrieve domain specific knowledge from medical textbooks corpus given a query. & \texttt{query} \\
 & \texttt{retrieve\_statpearls} & Retrieve clinical decision support from StatPearls corpus given a query. & \texttt{query} \\
 & \texttt{retrieve\_wikipedia} & Retrieve general knowledge from Wikipedia corpus given a query. & \texttt{query} \\
\bottomrule
\end{tabular}%
}
\caption{Overview of the Toolbox. The toolbox includes categories for Candidate extraction, Inner reasoning, Knowledge retrieval, Record querying, and Table schema inspection.}
\label{tab: Toolbox Overview}
\end{table*}

\section{Experimental Setting}
\label{appendix: experimental setting}

\subsection{Baselines}
To comprehensively verify the effectiveness and robustness of \agentname, we conduct comparative evaluations against a diverse set of state-of-the-art agent-based methods across five distinct LLM backbones. We employ a mix of powerful open-weights and proprietary models as backbones to ensure broad adaptability, including \textbf{Qwen3-30B-A3B}, \textbf{Qwen3-235B-A22B}, \textbf{Qwen3-Next-80B}~\cite{DBLP:journals/corr/abs-2505-09388}, \textbf{GPT-5-min}~\cite{openai2024gpt5}, and \textbf{Grok-4.1-fast}~\cite{xai2025grok41fast}.

On these backbones, we evaluate six representative agent methods as baselines, categorized into static and evolving methods. Static agents include standard \textbf{ReAct}~\cite{DBLP:conf/iclr/YaoZYDSN023}, self-reflecting \textbf{Reflexion}~\cite{DBLP:conf/nips/ShinnCGNY23}, and the unidirectional summarization method \textbf{ReSum}~\cite{DBLP:journals/corr/abs-2509-13313}. Evolving agents include \textbf{Reasoning Bank}~\cite{DBLP:journals/corr/abs-2509-25140} and \textbf{ReflecTool}~\cite{DBLP:conf/acl/LiaoJW025}, for which we evaluate both the \textbf{Candidate Selection} and \textbf{Iterative Refinement} variants.

\subsection{Implementation Details}
In our experiments, we set the maximum interaction turns for all agents to 100 to ensure computational feasibility while allowing sufficient exploration. The maximum context length is capped at 64,000 tokens to accommodate the potentially extensive history in EHR tasks. For evolving methods, we provide 100 training examples per task sampled from the common set of MIMIC-IV for experience accumulation. Crucially, we utilize this identical set of accumulated experiences to evaluate performance across all three datasets (\textit{MIMIC-IV-Common}, \textit{MIMIC-IV-Rare}, and \textit{MIMIC-III}), thereby strictly testing the cross-distribution generalization of the evolved agents. During inference, we retrieve only the top-1 most similar experience sample based on the embedding generated by \texttt{bge-m3}~\cite{bge-m3}. For \agentname, the retrospective summarization interval $w$ is set to 10 turns. All experiments are conducted on a cluster of NVIDIA 8xA100 GPUs. Finally, Prompt~\ref{prompt: diagnoses} to \ref{prompt: transfer} contains the instructions of six prompts in \benchname and Prompt~\ref{prompt: summarization} to \ref{prompt: summarizer experience generation} contains the prompt used in \agentname method.

\subsection{Metrics}
\label{appendix: metrics}
\paragraph{F1 Scores}
For a given clinical task instance, let $Y$ denote the set of elements predicted by the agent and $Y^*$ denote the set of ground truth elements. We first evaluate the Precision and Recall with the formula below:
\begin{align}
    \text{Precision} = \frac{|Y \cap Y^*|}{|Y|}, \\ 
    \text{Recall} = \frac{|Y \cap Y^*|}{|Y^*|}
\end{align}
The F1 Score is defined as the harmonic mean of Precision and Recall, providing a balanced single-value metric that penalizes both missed relevant items and incorrect predictions:
\begin{align}
    \text{F1} = \frac{2 \times \text{Precision} \times \text{Recall}}{\text{Precision} + \text{Recall}}
\end{align}

\paragraph{Best@K F1}
To quantify the expected peak performance given a specific inference budget $K$, we calculate the Best@K F1 Score based on a total pool of $N$ generated trajectories (where $N \ge K$). This metric is defined as the expected maximum F1 score when $K$ trajectories are sampled without replacement from the total pool $N$.For a specific instance $i$, let $\mathcal{T}_i = \{\tau_1, \tau_2, \dots, \tau_N\}$ be the set of all $N$ generated trajectories. Let $\mathcal{C}_K$ be the set of all possible subsets of $\mathcal{T}_i$ with size $K$, where the total number of such subsets is the binomial coefficient $\binom{N}{K}$. The Best@K score, for instance $i$ is calculated by averaging the maximum F1 score over all possible combinations:

\begin{equation}
\text{Best}@K \text{ F1} = \frac{1}{\binom{N}{K}} \sum_{S \in \mathcal{C}_K} \max_{\tau \in S} F1(\tau)
\end{equation}

This probabilistic definition encompasses two critical boundary conditions that provide insight into the model's behavior: 
(1) \textbf{Expected Performance ($K=1$):} When the budget is a single attempt, the metric collapses to the arithmetic mean of all generated trajectories, reflecting the model's average performance without selection:

\begin{equation}
\text{Best}@1 \text{ F1} = \frac{1}{\binom{N}{1}} \sum_{j=1}^{N} F1(\tau_j) = \frac{1}{N} \sum_{j=1}^{N} F1(\tau_j)\end{equation} 
(2) \textbf{Peak Potential ($K=N$):} When the budget allows evaluating the entire generated pool, the metric represents the absolute upper bound of the model's capability (Pass@N):
\begin{equation}
\text{Best}@N \text{ F1} = \frac{1}{\binom{N}{N}} \max_{\tau \in \mathcal{T}_i} F1(\tau) = \max_{j \in \{1, \dots, N\}} F1(\tau_j)
\end{equation}

This rigorous definition eliminates the variance associated with random sampling and ensures a deterministic evaluation of the trade-off between computational budget and performance.

\begin{figure*}[t]
    \centering
    \includegraphics[width=\linewidth]{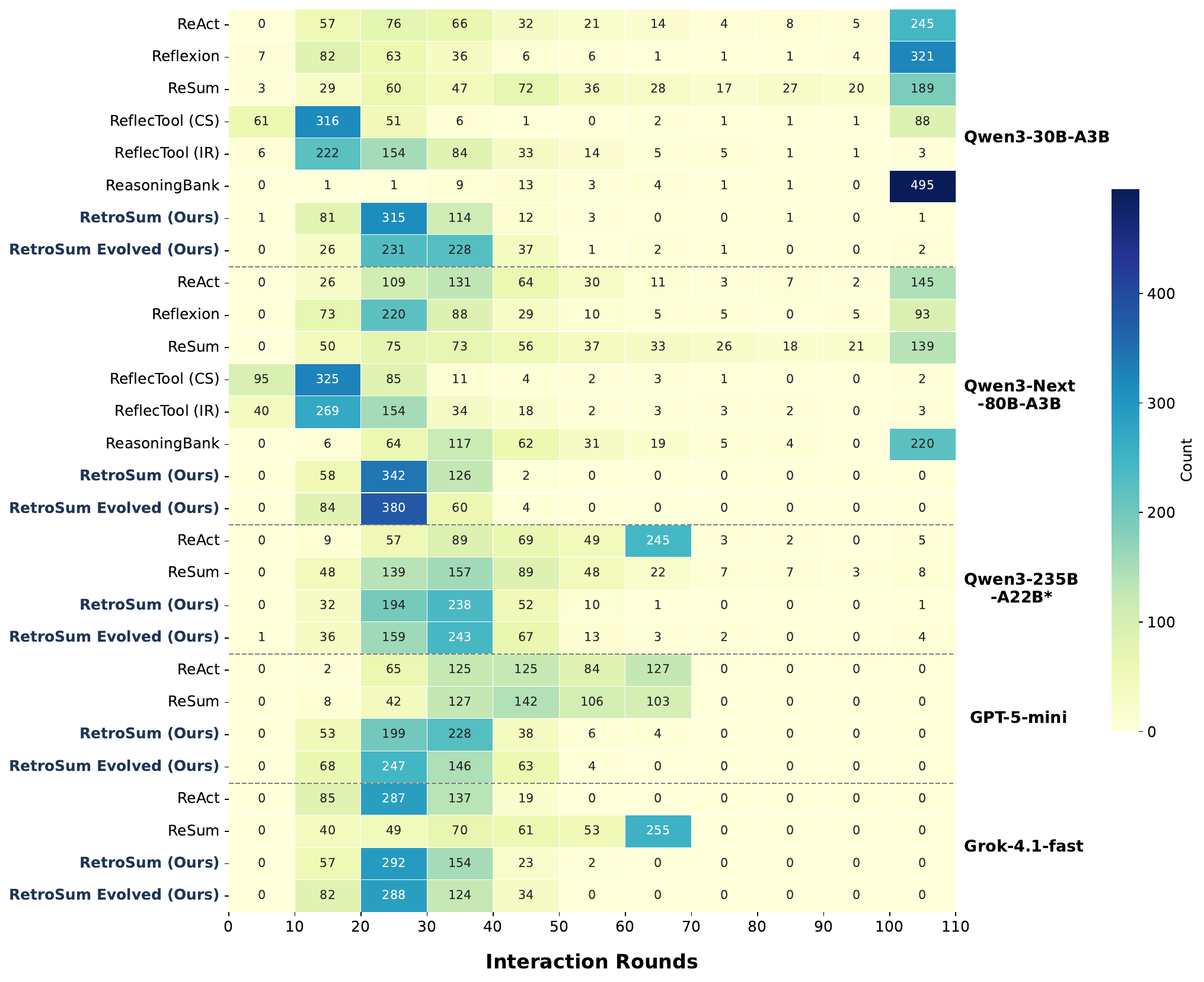}
    \caption{Heatmap visualization of interaction round distributions across different agent methods and backbones on the diagnoses task in \textbf{MIMIC-IV Common}. The color intensity denotes the frequency of cases falling within each interval.}
    \label{fig: Turn Heatmap}
\end{figure*}

\section{Error Definition}
\label{appendix: error_definition}

To systematically diagnose the pathological behaviors of the agents, we categorized the failure trajectories into six distinct types. These categories range from low-level format violations to high-level reasoning stagnation.

\paragraph{No Prediction.}
This category aggregates scenarios where the agent terminates the interaction episode without yielding a parsable or valid prediction. It encompasses four specific failure modes driven by structural or cognitive deficiencies. The first involves \textit{Tool Parsing Failure}, where the agent attempts to generate a tool invocation but fails to adhere to the required syntax, causing the system to misinterpret the action as a termination signal, which is exemplified in Case \ref{case: no_prediction}. The second includes \textit{Format Submersion}, where the agent successfully invokes the finish action and provides a text response, yet the answer extractor fails to parse a valid prediction from the unstructured output. The third mode is \textit{Answer Induced Failure}, occurring when the agent invokes the finish tool but provides an empty or null observation, resulting in a void prediction. Finally, the category includes \textit{Incomplete Termination}, where the interaction concludes—often due to reaching the maximum turn limit—without the agent ever invoking the definitive \texttt{finish} tool, thereby leaving the task unresolved.

\paragraph{Tool Repeat.}
We define this error as a state of exact cognitive stagnation. It is characterized by the agent executing the identical tool with completely identical parameters for five consecutive interaction turns, which is illustrated in Case \ref{case: tool_repeat}. This behavior indicates that the agent has fallen into a rigid loop, repeatedly querying the same information without updating its internal strategy, leading to severe information redundancy.

\paragraph{Single-Tool Loop.}
Distinct from exact repetition, this category characterizes scenarios where the agent resorts to a naive, iterative retrieval strategy by executing the same tool with negligible parameter variations, as shown in Case \ref{case: single_tool_loop}. We employ the Ratcliff-Obershelp algorithm to quantify the similarity between the parameters of consecutive tool calls sharing the same tool name. The string similarity score $S_{ro}$ between two parameter strings $P_1$ and $P_2$ is calculated as:
\begin{equation}
    S_{ro}(P_1, P_2) = \frac{2 \cdot K_m}{|P_1| + |P_2|}
\end{equation}
where $K_m$ represents the number of matching characters derived from the longest common subsequence and its recursive sub-segments. If the agent executes the same tool for ten consecutive rounds with a parameter similarity $S_{ro} > 0.95$, the trajectory is classified as a Single-Tool Loop. This behavior indicates that the agent is adopting a direct yet ineffective sequential scanning approach, which drastically degrades reasoning efficiency and inundates the context with excessive volumes of invalid or redundant query information.

\paragraph{Multi-Tool Cyclic Loop.}
This category identifies non-consecutive but pervasive repetition throughout the entire trajectory. Utilizing the same similarity metric $S_{ro}$ and threshold ($>0.95$), we flag a trajectory as a cyclic loop if similar tool calls appear more than 15 times in total across the interaction history, which is demonstrated in Case \ref{case: multi_tool_cyclic_loop}. This behavior suggests a lack of long-term planning, where the agent repeatedly revisits previously explored states or queries in a circular manner.

\paragraph{Tool Usage Error.}
This error type reflects hallucination or schema violation regarding the available toolbox. It occurs when the agent attempts to invoke a tool that does not exist in the defined toolbox or provides arguments that do not align with the tool's required parameter schema, as shown in Case \ref{case: tool_usage_error}. Such failures indicate a disconnect between the agent's reasoning core and the environmental constraints.

\paragraph{No Candidate Tool.}
Given the benchmark's design, valid predictions must be grounded in the candidate search space. This error is assigned when the agent fails to invoke any candidate-retrieval related tools (e.g., \texttt{get\_candidates\_by\_keyword}) throughout the entire interaction. The absence of such calls guarantees that the final answer will not map to a valid entry in the candidate tables, inevitably leading to a task failure, which can be seen in see Case \ref{case: no_candidate_tool}.

\begin{figure*}[t]
    \centering
    \includegraphics[width=\linewidth]{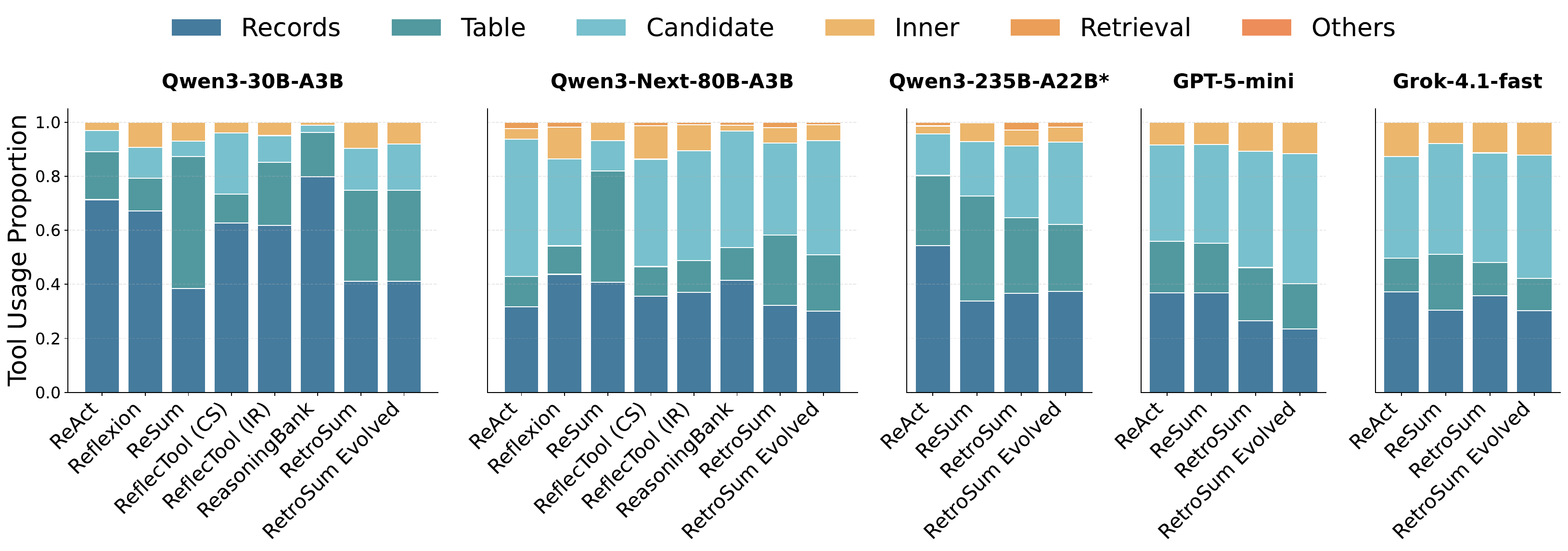}
    \caption{Proportional distribution of tool category usage across different agent methods and LLM backbones on the diagnoses task in \textbf{MIMIC-IV Common}.}
    \label{fig: Tool Stack Chart}
\end{figure*}

\section{Additional Experiments}
\label{appendix: experiments}

\subsection{Turn Analysis}
\label{appendix: turn analysis}
Following the specific evaluation of the Qwen3-30B-MoE architecture in Section \ref{subsec: Turn Analysis}, we extend our investigation using the interaction turn heatmap illustrated in Figure \ref{fig: Turn Heatmap} to examine the broader impact of varying model capabilities and agent methods.

The visualization exposes that baseline approaches exhibit significant sensitivity to the underlying model architecture. Methods such as ReAct and ReasoningBank demonstrate high volatility across different backbones. On weaker models, ReasoningBank specifically suffers from catastrophic stagnation where nearly 500 cases reach the maximum turn limit. This phenomenon indicates that in the absence of an effective memory mechanism, lossy context compression leads to the omission of critical details. Consequently, the agent is forced into redundant information-seeking loops to attempt to recover this lost context. In contrast, \agentname maintains a consistent distribution of interaction rounds regardless of the backbone model scale. This stability suggests that our retrospective mechanism functions as a regularizer which self-standardizes the reasoning trajectory and effectively enhances reasoning efficiency.

Conversely, distinct failure modes are observed in alternative self-evolving methods such as ReflecTool, which show a tendency toward premature termination. These agents frequently conclude episodes within 10 to 20 rounds, evidenced by ReflecTool-CS resolving 316 cases within this short interval on Qwen3-30B-A3B. In the context of complex clinical diagnostics, such rapid convergence often implies superficial reasoning or a failure to conduct necessary differential diagnosis verifications.

\subsection{Tool Analysis}

To investigate the behavioral distinctness of \agentname, we analyzed the distribution of tool calls across different backbones, as depicted in Figure \ref{fig: Tool Stack Chart}. This breakdown reveals fundamental differences in how agents navigate the clinical decision-making process.

\paragraph{Behavior with Different Methods.}
A prominent trend observed in baseline methods, particularly ReAct and Reflexion, is the dominance of the \textbf{Records} tool. On weaker backbones like Qwen3-30B-A3B, these methods dedicate over 60\% of their actions to fetching raw EHR records. This pattern indicates an inefficient information acquisition strategy, where the agent struggles to locate relevant evidence and repeatedly queries the database. 
In contrast, \agentname exhibits a significantly more balanced distribution. Notably, the proportion of Candidate tool usage is consistently higher in \agentname compared to baselines. This shift suggests that our retrospective mechanism successfully compresses the context, enabling the agent to progress beyond superficial data acquisition. Instead, the agent prioritizes the more advanced stage of active grounding, where it systematically aligns clinical findings against the candidate space to refine its predictions.

\paragraph{Behavioral Alignment with Stronger Models.}
The tool usage patterns also highlight how \agentname bridges the capability gap between models. As we observe the transition from smaller models to more powerful reasoning models, there is a natural tendency for agents to increase their usage of \textbf{Candidate} and \textbf{Inner} tools, reflecting a more purposeful reasoning process. Crucially, \agentname enables smaller models to mimic this "expert" behavior pattern, exhibiting a tool distribution profile that resembles that of GPT-5-mini baselines. This indicates that our framework effectively guides weaker models to adopt the efficient behavioral heuristics naturally found in stronger foundation models.

\paragraph{Schema Awareness.}
Furthermore, the usage of the \textbf{Table} tool provides insight into schema understanding. While some baselines fluctuate in their reliance on schema inspection, \agentname maintains a steady but moderate usage. This implies that the agent retains sufficient awareness of the database structure through its memory mechanism, reducing the need for redundant schema queries while ensuring precise SQL generation for record retrieval.

\begin{figure*}[t]
    \centering
    \includegraphics[width=\linewidth]{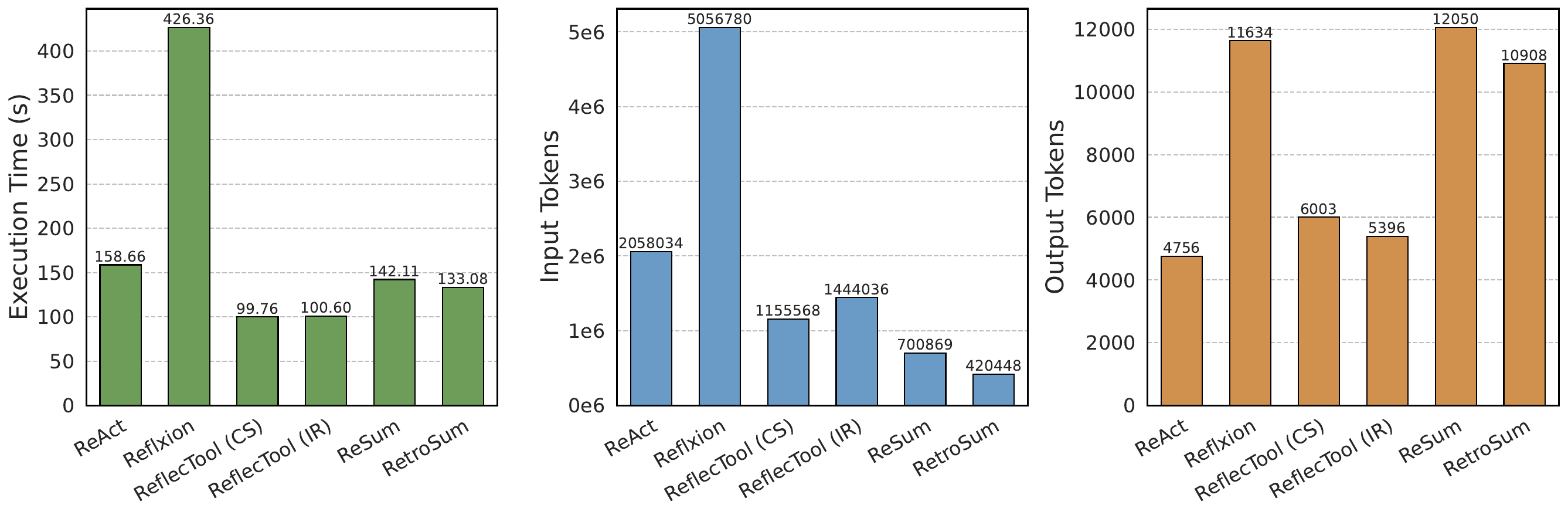}
    \caption{Computational cost analysis across different agent frameworks. We report the average execution time per sample (left), average input tokens per sample (middle), and average output tokens per sample (right). \agentname demonstrates superior efficiency, achieving the lowest input consumption (~4.9$\times$ reduction vs. ReAct) while also reducing execution latency compared to standard baselines like ReSum and ReAct.}
    \label{fig: computational consumption}
\end{figure*}

\subsection{Computational Consumption Analysis}
\label{appendix: computational consumption analysis}
Beyond predictive performance, the deployment of agents in real-world clinical settings requires a rigorous assessment of computational costs and latency. Figure \ref{fig: computational consumption} illustrates the resource consumption of \agentname compared to baseline methods across three key metrics: execution time, input tokens, and output tokens.

\paragraph{Token Economy} \agentname demonstrates a remarkable advantage in input token efficiency. By dynamically compressing the interaction history into concise retrospective summaries, our method reduces the average input tokens per sample to approximately 0.42M, representing a 4.9$\times$ reduction compared to ReAct (2.06M) and a 1.7$\times$ reduction compared to ReSum (0.70M). This drastic decrease is critical for cost-sensitive applications, as input tokens typically dominate the inference cost of Large Language Models. Conversely, while \agentname generates an average of 10.9k output tokens—comparable to ReSum (12.1k)—this "investment" in detailed reasoning is significantly outweighed by the massive savings in input context processing.

\paragraph{Execution Latency} Despite the additional computational overhead required for generating summaries and evolving experiences, \agentname achieves an average execution time of 133.08s, which is notably faster than both ReSum (142.11s) and ReAct (158.66s). This result indicates that the retrospective mechanism successfully optimizes the overall workflow—by preventing the agent from getting lost in long, repetitive raw contexts (as seen in ReAct), \agentname reduces the total number of interaction turns, thereby achieving a superior trade-off that lowers both latency and cost.

\subsection{Record Retrieved Distribution}
\begin{landscape}
\begin{figure}[t]
    \centering
    \includegraphics[width=1.5\textwidth]{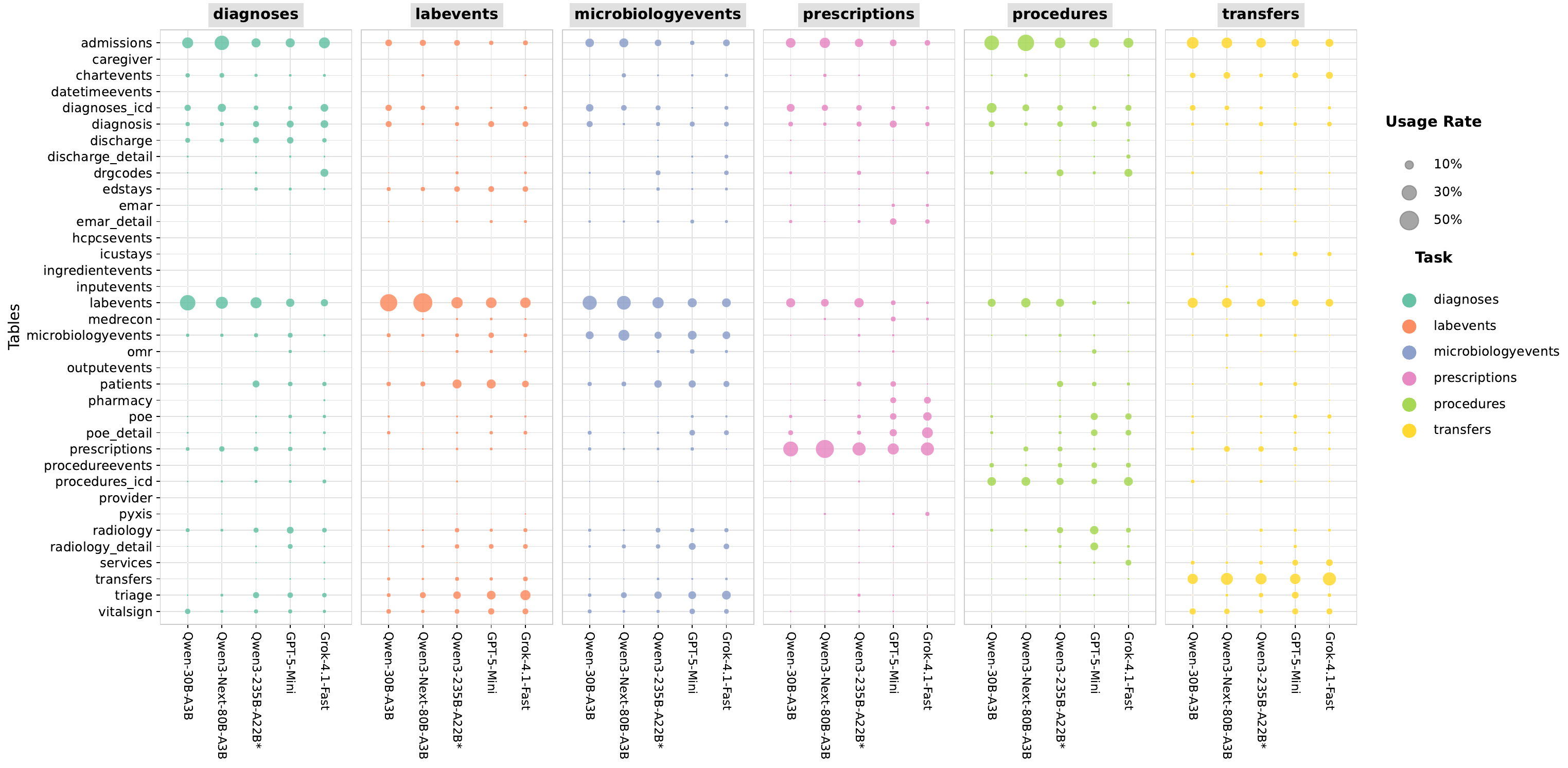}
    \caption{Distribution of retrieved EHR tables aggregated by different \textbf{base models}. The visualization averages table access frequencies across all agent methods to highlight the intrinsic retrieval preferences of each LLM. Panels are segmented by task, with dot size representing the usage rate of specific tables by each model.}
    \label{fig:Table Heatmap Models}
\end{figure}
\end{landscape}

\begin{landscape}
\begin{figure}[t]
    \centering
    \includegraphics[width=\linewidth]{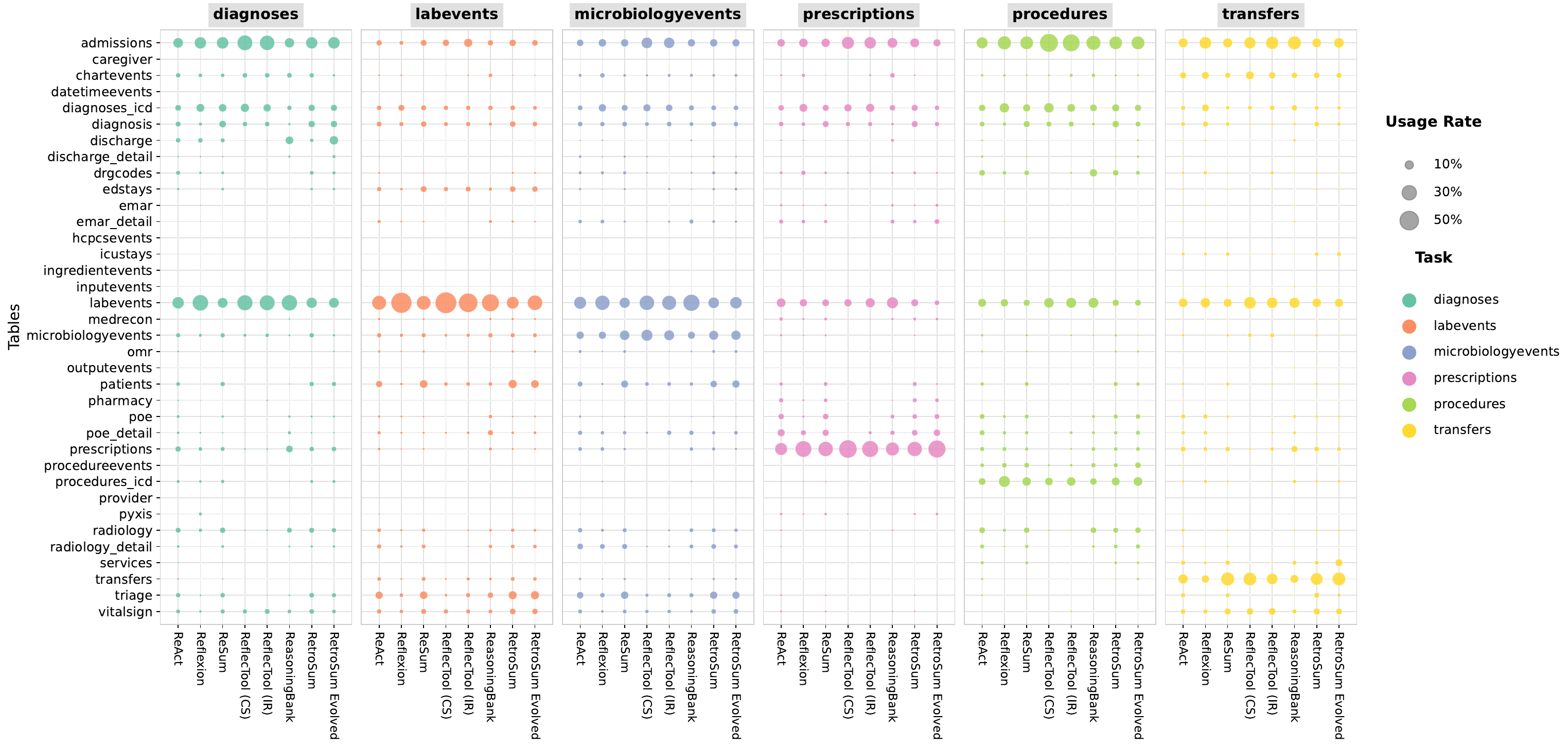}
    \caption{Distribution of retrieved EHR tables aggregated by different \textbf{agent methods}. The visualization averages table access frequencies across all backbone models to isolate the specific retrieval strategies of each method. Panels are segmented by task, with dot size representing the usage rate of specific tables by each agent framework.}
    \label{fig: Table Heatmap Methods}
\end{figure}
\end{landscape}

To elucidate the information-seeking dynamics of different agents, we analyze the distribution of retrieved tables across various clinical tasks. Figure \ref{fig:Table Heatmap Models} and Figure \ref{fig: Table Heatmap Methods} visualize the access frequency of EHR tables, aggregated by backbone model and agent framework, respectively.

\paragraph{Task-Specific Schema Alignment.}
A clear pattern observed across all heatmaps is the strong semantic alignment between the task domain and the retrieved tables. As evidenced in the visualizations, agents predominantly query tables that are intrinsically relevant to the target task. For instance, in the \texttt{labevents} task, the retrieval focus is heavily concentrated on the \texttt{labevents} table. This dominant diagonal pattern confirms that the agents possess a fundamental awareness of the database schema and can identify the primary information sources required for specific clinical queries.

\paragraph{Impact of Model Capabilities.}
The aggregation by backbone model reveals distinct retrieval behaviors correlated with model scale. Smaller models, such as the Qwen3-30B-A3B and Qwen3-80B-A3B, exhibit a pattern of repetitive intensity, concentrating heavily on a limited set of tables. This suggests a lower efficiency in information extraction, necessitating repeated access to grasp the context. In contrast, stronger models like GPT-5-mini and Grok-4.1 display a more distributed attention mechanism, synthesizing information from a wider array of sources rather than fixating on a single table. 
Specific table preferences further highlight these differences. While all models prioritize \texttt{labevents}, which stores all lab test results including hematology and chemistry, the Qwen series accesses it with excessive frequency compared to the more balanced usage by GPT-5-mini and Grok-4.1. Additionally, Grok-4.1 demonstrates a unique preference for the \texttt{triage} table, which contains initial emergency department assessment data and vital signs, whereas smaller models rely more heavily on the basic \texttt{admissions} table. In summary, advanced models demonstrate a capacity for holistic clinical contextualization, utilizing peripheral data sources like triage to inform their reasoning, while smaller models adhere to a more rigid and repetitive focus on core administrative tables.

\paragraph{Strategies of Agent Methods.}
Disaggregating the results by framework reveals how different mechanisms influence information gathering. Methods incorporating summarization capabilities, such as ReSum and \agentname, exhibit a significantly broader and more uniform query distribution across tables. This indicates that the summarization module facilitates the efficient digestion of table content, encouraging the agent to explore diverse information sources to construct a comprehensive patient profile. Conversely, methods incorporating self-evolution and self-reflection, such as Reflexion, ReflecTool, and \agentname Evolved, demonstrate a sharper focus. These methods leverage past paradigms to identify and concentrate on the most relevant tables for the specific task. Notably, our approach integrates the advantages of both categories. It supports active exploration to acquire broad context while maintaining a task-oriented focus on critical data, preventing the agent from drifting into irrelevant information.

\paragraph{Task-Dependent Retrieval Scope.}
From a task perspective, the retrieval patterns validate the multi-dimensional design of our benchmark. We observe a clear distinction between specialized and foundational clinical tasks. Tables associated with \texttt{prescriptions}, \texttt{procedures}, and \texttt{transfers} are accessed almost exclusively during their respective tasks, indicating a high degree of specificity. In contrast, \texttt{diagnoses}, \texttt{labevents}, and \texttt{microbiologyevents} serve as foundational patient information sources and are utilized across a wider range of scenarios. This implies that the first three tasks evaluate the agent's ability to handle specialized, domain-specific queries, while the latter three assess the ability to synthesize comprehensive clinical subjects. This dichotomy ensures that our benchmark provides a robust evaluation of clinical capabilities across multiple dimensions of complexity and specificity.

\subsection{Case Study}
To provide a concrete understanding of \agentname's operational logic, we present qualitative examples of its reasoning process and evolving mechanisms. Please note that due to the excessive length of raw EHR observation logs, we employed an LLM to condense the interaction history in these illustrations, preserving the core reasoning logic while ensuring readability.

We first present a complete inference process in Case~\ref{case: trajectory}, demonstrating how \agentname navigates the complex EHR environment to conclude a clinical task. Furthermore, to isolate the impact of our evolving strategy, we provide specific instances of experience retrieval: Case~\ref{case: actor experience} showcases the application of historical insights by the Actor module, while Case~\ref{case: summarizer experience} illustrates the effect of retrieved experiences on the Summarizer module.

\section{AI Assistance Statement}
Language editing and stylistic refinement of the draft were performed with the aid of large language models, including ChatGPT~\footnote{\url{https://chatgpt.com/}} and Google Gemini~\footnote{\url{https://gemini.google.com/app/}}.

\begin{prompt}
\label{prompt: diagnoses}
\textbf{Diagnose Task Prompt} \newline 
Your current task is to act as a diagnostician. \newline 
\newline 
Your objective is to determine all plausible diagnoses for the patient's current condition by analyzing the patient's complete history. \newline 
\newline 
You must find the most likely official CCS candidates using the **diagnoses\_ccs\_candidates** reference table.\newline 
\newline 
Present your final answer as a **list format** with `finish` tool calling, which must contain **multiple plausible diagnoses**. Each item in the list must be a string representing an official CCS diagnosis name, and **must not contain any codes or other additional information**.
\end{prompt}

\begin{prompt}
\textbf{Labevents Task Prompt} \newline 
Your current task is to act as a laboratory medicine specialist. \newline 
\newline 
Your objective is to determine all necessary laboratory tests for the patient by analyzing their complete medical history, current clinical condition, and established diagnoses.\newline 
\newline 
You should provide as many laboratory tests as possible to cover the patient's current clinical condition.\newline 
\newline 
You must find the most likely official laboratory test candidates using the **labevents\_candidates** reference table.\newline 
\newline 
Present your final answer as a **list format** with `finish` tool calling, which must contain **multiple plausible laboratory tests**. Each item in the list must be a string representing an official laboratory test name, and **must not contain any codes or other additional information**.
\end{prompt}

\begin{prompt}
\textbf{Microbiology Task Prompt} \newline 
Your current task is to act as a clinical microbiologist.\newline 
\newline 
Your objective is to determine all necessary microbiological tests for the patient by analyzing their complete medical history, current clinical condition, established diagnoses, and clinical signs of infection.\newline 
\newline 
You must find the most likely official microbiological test candidates using the **microbiologyevents\_candidates** reference data or semantic matching tools.\newline 
\newline 
Present your final answer as a **list format** with `finish` tool calling, which must contain **multiple plausible microbiological tests**. Each item in the list must be a string representing an official microbiological test name, and **must not contain any codes or other additional information**.
\end{prompt}

\begin{prompt}
\textbf{Prescriptions Task Prompt} \newline 
Your current task is to act as a pharmacist.\newline 
\newline 
Your objective is to determine all necessary ATC therapeutic categories for the patient by analyzing their complete medical history, current clinical condition, and established diagnoses.\newline 
\newline 
You must find the most likely official ATC name candidates using the **prescriptions\_atc\_candidates** reference data or semantic matching tools.\newline 
\newline 
Present your final answer as a **list format** with `finish` tool calling, which must contain **multiple plausible ATC names**. Each item in the list must be a string representing an official ATC name, and **must not contain any codes or other additional information**.
\end{prompt}

\begin{prompt}
\textbf{Procedures Task Prompt} \newline 
Your current task is to act as a surgical planner.\newline 
\newline 
Your objective is to determine all necessary surgical procedures for the patient by analyzing their complete medical history and established diagnoses.\newline 
\newline 
You must find the most likely official CCS procedure candidates using the **`procedures\_ccs\_candidates`** reference table.
\newline 
Present your final answer as a **list format** with `finish` tool calling, which must contain **multiple plausible procedures**. Each item in the list must be a string representing an official CCS procedure name, and **must not contain any codes or other additional information**.
\end{prompt}

\begin{prompt}
\label{prompt: transfer}
\textbf{Transfer Task Prompt} \newline 
Your current task is to act as a hospital care coordinator and clinical decision-maker.\newline 
\newline 
Your objective is to determine the most appropriate care unit for patient transfer by analyzing their current clinical condition, medical history, severity of illness, and care requirements.\newline 
\newline 
You must consider the patient's current location, clinical stability, required level of monitoring, and specialized care needs to recommend the optimal transfer destination.\newline 
\newline 
You must find the most likely official care unit candidates using the **transfers\_candidates** reference data or semantic matching tools.\newline 
\newline 
Present your final answer as a **list format** with `finish` tool calling, which must contain **multiple plausible care units**. Each item in the list must be a string representing an official care unit name, and **must not contain any codes or other additional information**.
\end{prompt}

\begin{prompt}
\label{prompt: summarization}
\textbf{Summarization  Prompt} \newline 
You are an expert at analyzing conversation history and extracting relevant information. Your task is to thoroughly evaluate the conversation history and current question to provide a comprehensive summary that will help solve the task. \newline 
 \newline 
\#\# Task Guidelines  \newline 
1. Information Analysis: \newline 
   - Carefully analyze the conversation history to identify truly useful information. \newline 
   - Focus on information that directly contributes to answering the question. \newline 
   - Do NOT make assumptions, guesses, or inferences beyond what is explicitly stated in the conversation. \newline 
   - If information is missing or unclear, do NOT include it in your summary. \newline 
 \newline 
2. Summary Requirements: \newline 
   - Extract only the most relevant information that is explicitly present in the conversation. \newline 
   - Synthesize information from multiple exchanges when relevant. \newline 
   - Only include information that is certain and clearly stated in the conversation. \newline 
   - Do NOT output or mention any information that is uncertain, insufficient, or cannot be confirmed from the conversation. \newline 
 \newline 
3. Output Format: Your response should be structured as follows: \newline 
<summary> \newline 
- Essential Information: [Organize the relevant and certain information from the conversation history that helps address the question.] \newline 
</summary> \newline 
 \newline 
Strictly avoid fabricating, inferring, or exaggerating any information not present in the conversation. Only output information that is certain and explicitly stated. \newline 
 \newline 
Question \newline 
\{question\}  \newline 
 \newline 
Conversation History \newline 
\{recent\_history\_messages\} \newline 
 \newline 
Please generate a comprehensive and useful summary. Note that you are not permitted to invoke tools during this process.
\end{prompt}

\begin{prompt}
\textbf{Actor Experiences Generation Prompt} \newline 
\small
You are an expert in clinical reasoning auditor. You will be provided with a complete post-hoc analysis package of a clinical reasoning task performed by an AI Actor agent.\newline 
\newline 
Your inputs include:\newline 
1.  **User Query**: The original clinical question or task.\newline 
2.  **Prediction Result \& Ground Truth**: What the Actor ultimately concluded versus the correct answer.\newline 
4.  **Complete Raw Trajectory / Actions Taken**: The exact actions the Actor took based on the summaries.\newline 
\newline 
\#\# Guidelines\newline 
Your task is to analyze the **reasoning quality** of the Actor agent. You need to extract useful insights in the format of memory items that focus on improving future clinical decision-making processes. The goal is to identify where the Actor's logic was flawed, overly cautious, too aggressive.\newline 
\newline 
\#\# Important notes\newline 
- **Focus on the Actor's Decisions:** Focus on: Given the information *present* at step T, did the Actor make the most logical action?\newline 
- **Analyze Reasoning Gaps:**\newline 
    - If failed: Did the Actor jump to a conclusion not supported by the retrieved information? Did it ignore conflicting evidence presented in the EHR records? Did it fail to order a necessary confirmatory test suggested by the recprds ambiguity?\newline 
    - If successful: What robust reasoning strategy did the Actor use to navigate uncertainty or complex data presented in the summaries?\newline 
- **Generalizable Reasoning Principles:** Insights should be about *how to think* clinically (e.g., differential diagnosis strategies, handling conflicting data, recognizing urgency), not about specific medical facts.\newline 
    - Do not summarize strategies that are already mentioned in the raw instructions; focus on the implicit reasoning strategies.\newline 
    - You can extract at most {max\_items} memory items.\newline 
    - You must not repeat similar or overlapping items.\newline 
\newline 
\#\# Output Format\newline 
Your output must strictly follow the Markdown format shown below. Ensure ALL fields (Title, Description, Content) are provided.\newline 
\newline 
Required Format:\newline 
''' \newline 
\# Memory Item i\newline 
\#\# Title <short title, max 15 words>\newline 
\#\# Description <one sentence summary of the memory item> \newline 
\#\# Content <1-3 sentences describing the insights learned to successfully accomplish the task> \newline 
''' \newline 
\newline 
\# Query:\newline 
\{query\}\newline 
\newline 
\# Complete Trajectory: \newline 
\{raw\_trajectory\}\newline 
\newline 
\# Prediction Result: \newline 
\{prediction\_result\}\newline 
\newline 
\# Ground Truth:\newline  
\{ground\_truth\}
\end{prompt}

\begin{prompt}
\label{prompt: summarizer experience generation}
\textbf{Summarizer Experiences Generation Prompt} \newline 
\small
You are an expert medical data summarization auditor. You will be provided with a complete post-hoc analysis package of a clinical reasoning task.\newline 
\newline 
Your inputs include:\newline 
1.  **User Query**: The original clinical question or task.\newline 
2.  **Prediction Result \& Ground Truth**: Knowing the final outcome is crucial for Hindsight bias analysis.\newline 
3.  **Complete Raw Trajectory**: The exhaustive, detailed truth (e.g., full EHR tables).\newline 
4.  **Summarized Trajectory**: The compressed version generated by the Summarizer.\newline 
\newline 
\#\# Guidelines\newline 
Your task is to conduct a **hindsight gap analysis** between the raw data and the generated summaries. You need to extract insights to improve how future Summarizers extract and compress clinical information.\newline 
\newline 
\#\# Important notes \newline 
- **Hindsight Gap Analysis (Crucial):** You must use the `Ground Truth` to identify what was truly important in the `Raw Trajectory`. Then, check if that important information was captured accurately in the `Summarized Trajectory`.\newline 
- **Focus on Information Loss/Distortion:**\newline 
    - **What was lost?** Identify critical raw data points (e.g., a subtle lab trend, a specific timing of medication) that were essential for the correct diagnosis but were omitted or overly generalized in the summary.\newline 
    - **Why it matters?** Explain how omitting this specific type of raw data leads to downstream reasoning errors.\newline 
- **Generalizable Summarization Principles:** Insights should be rules for *how to summarize* better (e.g., "Always preserve exact values for abnormal vitals instead of just stating 'abnormal'", "Explicitly link temporal relationships between medication administration and subsequent lab changes").\newline 
- You can extract at most {max\_items} memory items.\newline 
- You must not repeat similar or overlapping items.\newline 
\newline 
\#\# Output Format \newline 
Your output must strictly follow the Markdown format shown below. Ensure ALL fields (Title, Description, Content) are provided.\newline 
\newline 
Required Format:\newline 
'''\newline 
\# Memory Item i\newline 
\#\# Title <short title, max 15 words>\newline 
\#\# Description <one sentence summary of the memory item>\newline 
\#\# Content <1-3 sentences describing the insights learned to successfully accomplishing the task>\newline 
'''\newline 
\newline 
\# Query:\newline 
\{query\}\newline 
\newline 
\# Complete Trajectory: \newline 
\{raw\_trajectory\}\newline 
\newline 
\# Summarized Trajectory: \newline 
\{summarized\_trajectory\}\newline 
\newline 
\# Prediction Result: \newline 
\{prediction\_result\}\newline 
\newline 
\# Ground Truth: \newline 
\{ground\_truth\}\newline 
\end{prompt}

\begin{case}
\label{case: no_prediction}
\textbf{No Prediction (Tool Parsing Failure)} \newline 
... (Previous steps omitted) \newline 
\newline 
\#\# Step 63:\newline 
- \textbf{Action}: finish\{'response': '<tool\_call>\textbackslash n\{"name": "get\_candidates\_by\_fuzzy\_matching", "arguments": \{"table\_name": "diagnoses\_ccs\_candidates", "keywords": ["myelodysplastic", "malignancy", "leukemia", ... \textit{[truncated 1000+ keywords]} ... "fraction", "EF", "'\} \newline 
- \textbf{Observation}: Task concluded. No valid prediction format detected in response. \newline 
\newline 
\textcolor{red}{\textbf{\# Error Analysis: The agent attempted to generate a tool invocation (`get\_candidates...`) but failed to adhere to the syntax, outputting it as a raw string within the `finish` action. The system misinterpreted this as a termination signal with invalid content, resulting in zero predictions.}}
\end{case}

\begin{case}
\label{case: tool_repeat}
\textbf{Tool Repeat Error} \newline 
... (Previous steps 1-20 omitted) \newline 
\newline 
\#\# Step 21:\newline 
- \textbf{Action}: get\_records\_by\_keyword\{'subject\_id': '19505049', 'table\_name': 'admissions', 'keyword': 'vomiting'\} \newline 
- \textbf{Observation}: No records found in table 'admissions' containing the keyword 'vomiting'. \newline 
\newline 
\#\# Step 22:\newline 
- \textbf{Action}: get\_records\_by\_keyword\{'subject\_id': '19505049', 'table\_name': 'admissions', 'keyword': 'vomiting'\} \newline 
- \textbf{Observation}: No records found in table 'admissions' containing the keyword 'vomiting'. \newline 
\newline 
\textit{... (Steps 23-29 repeat the exact same action and parameters) ...} \newline 
\newline 
\#\# Step 30:\newline 
- \textbf{Action}: get\_records\_by\_keyword\{'subject\_id': '19505049', 'table\_name': 'admissions', 'keyword': 'vomiting'\} \newline 
- \textbf{Observation}: No records found in table 'admissions' containing the keyword 'vomiting'. \newline 
\newline 
\textcolor{red}{\textbf{\# Error Analysis: The agent executes identical actions with identical parameters for $>$5 consecutive turns, failing to update its strategy despite negative feedback.}}
\end{case}

\begin{case}
\label{case: single_tool_loop}
\textbf{Single-Tool Loop} \newline 
... (Previous steps omitted) \newline 
\newline 
\#\# Step 19:\newline 
- \textbf{Action}: get\_records\_by\_value\{'subject\_id': ‘17332316’, 'table\_name': 'labevents', 'column\_name': 'itemid', 'value': '50889'\} \newline 
- \textbf{Observation}: No records found in table 'labevents' where 'itemid' equals '50889'. \newline 
\newline 
\#\# Step 20:\newline 
- \textbf{Action}: get\_records\_by\_value\{'subject\_id': ‘17332316’, 'table\_name': 'labevents', 'column\_name': 'itemid', 'value': '50894'\} \newline 
- \textbf{Observation}: No records found in table 'labevents' where 'itemid' equals '50894'. \newline 
\newline 
\#\# Step 21:\newline 
- \textbf{Action}: get\_records\_by\_value\{'subject\_id': ‘17332316’, 'table\_name': 'labevents', 'column\_name': 'itemid', 'value': '50883'\} \newline 
- \textbf{Observation}: labevent\_id ... 50883 ... Bilirubin, Direct ... [5 rows retrieved] \newline 
\newline 
\#\# Step 22:\newline 
- \textbf{Action}: get\_records\_by\_value\{'subject\_id': ‘17332316’, 'table\_name': 'labevents', 'column\_name': 'itemid', 'value': '50882'\} \newline 
- \textbf{Observation}: ... [Massive table data: Bicarbonate results, 16,427 chars truncated] ... \newline 
\newline 
\textit{... (Steps 23-28 continue querying 'itemid' with values 50881, 50884, 50885, 50890, etc.) ...} \newline 
\newline 
\textcolor{red}{\textbf{\# Error Analysis: The agent performs sequential scanning with high parameter similarity ($S_{ro} > 0.95$). Instead of filtering by time or flag, it blindly guesses IDs, inundating the context with excessive raw data while making negligible reasoning progress.}}
\end{case}

\begin{case}
\label{case: multi_tool_cyclic_loop}
\textbf{Multi-Tool Cyclic Loop} \newline 
... (Previous steps omitted) \newline 
\newline 
\textit{\textbf{--- Cycle Iteration 1 ---}} \newline 
\#\# Step 29:\newline 
- \textbf{Action}: get\_records\_by\_value\{'subject\_id': '12593903', 'table\_name': 'prescriptions', 'column\_name': 'drug', 'value': 'Sodium Chloride'\} \newline 
- \textbf{Observation}: No records found in table 'prescriptions' where 'drug' equals 'Sodium Chloride'. \newline 
\newline 
\#\# Step 30:\newline 
- \textbf{Action}: get\_records\_by\_value\{'subject\_id': '12593903', 'table\_name': 'prescriptions', 'column\_name': 'drug', 'value': 'D5 1/2NS'\} \newline 
- \textbf{Observation}: ... [Prescription found: 1000mL Bag D5 1/2NS ...] \newline 
\newline 
\textit{\textbf{--- Cycle Iteration 2 (Repeating steps 29-30) ---}} \newline 
\#\# Step 41:\newline 
- \textbf{Action}: get\_records\_by\_value\{'subject\_id': '12593903', 'table\_name': 'prescriptions', 'column\_name': 'drug', 'value': 'Sodium Chloride'\} \newline 
- \textbf{Observation}: No records found in table 'prescriptions' where 'drug' equals 'Sodium Chloride'. \newline 
\newline 
\#\# Step 42:\newline 
- \textbf{Action}: get\_records\_by\_value\{'subject\_id': '12593903', 'table\_name': 'prescriptions', 'column\_name': 'drug', 'value': 'D5 1/2NS'\} \newline 
- \textbf{Observation}: ... [Same prescription record retrieved again] \newline 
\newline 
\textit{... (Agent continues cycling through 'Carbohydrates', 'Glucose', '1/2 NS' repeatedly) ...} \newline 
\newline 
\textit{\textbf{--- Cycle Iteration N (Still stuck at end of trajectory) ---}} \newline 
\#\# Step 99:\newline 
- \textbf{Action}: get\_records\_by\_value\{'subject\_id': '12593903', 'table\_name': 'prescriptions', 'column\_name': 'drug', 'value': 'Sodium Chloride'\} \newline 
- \textbf{Observation}: No records found in table 'prescriptions' where 'drug' equals 'Sodium Chloride'. \newline 
\newline 
\textcolor{red}{\textbf{\# Error Analysis: The agent exhibits a non-consecutive but pervasive cyclic pattern. Similar tool calls ($S_{ro} > 0.95$) appear $>$15 times, indicating a failure in long-term planning and state tracking.}}
\end{case}

\begin{case}
\label{case: tool_usage_error}
\textbf{Tool Usage Error} \newline 
... (Previous steps omitted) \newline 
\newline 
\#\# Step 52:\newline 
- \textbf{Action}: get\_records\_by\_value\{'subject\_id': '13762777', 'table\_name': 'radiology\_detail', 'column\_name': 'hadm\_id', 'value': '22827736'\} \newline 
- \textbf{Observation}: Error: Column 'hadm\_id' not found in table 'radiology\_detail'. \newline 
\newline 
\textcolor{red}{\textbf{\# Error Analysis: The agent committed a schema violation by attempting to query the `radiology\_detail` table using a non-existent column `hadm\_id`. This indicates a failure to verify table metadata before execution, resulting in a runtime error.}}
\end{case}

\begin{case}
\label{case: no_candidate_tool}
\textbf{No Candidate Tool Failure} \newline 
... (Previous 98 steps omitted: The agent repeatedly used action 'get\_records\_by\_keyword') \newline 
\newline 
\#\# Step 99: \newline 
- \textbf{Action}: get\_records\_by\_keyword\{'subject\_id': '16002903', 'table\_name': 'radiology','keyword': 'postoperative'\} \newline 
- \textbf{Observation}: 'note\_id': '16002903-RR-10', 'subject\_id': '16002903', 'hadm\_id': '28925885.0', 'charttime': '2118-04-03 12:54:00', 'text': 'INDICATION: Status post right upper lobe wedge resection. COMPARISON: ... [remainder omitted]'\newline 
\newline 
\#\# Step 100: \newline 
- \textbf{Action}: get\_records\_by\_keyword\{'subject\_id': '16002903', 'table\_name': 'radiology', 'keyword': 'wedge resection'\} \newline 
- \textbf{Observation}: 'EXAMINATION: CHEST (PORTABLE AP) INDICATION: \_\_\_ year old woman with decline in resp function post-op R thoracotomy and wedge resection, now reintubated // Please eval interval change. COMPARISON: ... [remainder omitted]'\newline 
\newline 
\#\# Step 101: \newline 
- \textbf{Action}: finish\{'response': ['Lung cancer', 'Postoperative status after right upper lobe wedge resection', 'Pleural effusion', 'Pneumothorax', 'Chronic obstructive pulmonary disease']\} \newline 
- \textbf{Observation}: ['Lung cancer', 'Postoperative status after right upper lobe wedge resection', 'Pleural effusion', 'Pneumothorax', 'Chronic obstructive pulmonary disease']\newline 
\newline 
\textcolor{red}{\textbf{\# Error Analysis: The agent directly outputted free-text descriptions. It failed to invoke `get\_candidates\_by\_...` to map these findings to standardized CCS codes, rendering the prediction invalid.}}
\end{case}

\begin{case}
\label{case: actor experience}
\textbf{Actor Experience} \newline 
\#\# Prioritize High-Confidence Semantic Matches Over Fuzzy Search \newline 
- Description: When using semantic search, prioritize results with high similarity scores and validate against clinical plausibility rather than accepting low-similarity matches as valid. \newline 
- Content: The Actor accepted "Substance-related disorders" and "Headache; including migraine" based on moderate semantic similarity, but failed to recognize that the highest-scoring match for "barbiturate dependence" was "Alcohol-related disorders" (0.580), which is a more accurate and clinically relevant diagnosis. Relying on fuzzy or low-similarity matches can lead to misdiagnosis. \newline 
 \newline 
\#\# Distinguish Between Primary Diagnosis and Secondary Findings \newline 
- Description: Do not equate a positive drug screen (e.g., tricyclics) with a primary diagnosis of poisoning; consider the context of chronic use vs. acute toxicity. \newline 
- Content: The Actor listed "Poisoning by psychotropic agents" as a diagnosis based on a positive tricyclic screen, but this reflects medication use, not acute poisoning. In the absence of symptoms of overdose, this should be considered a finding, not a primary diagnosis, and should not be included in the final list of plausible conditions. \newline 
 \newline 
\#\# Avoid Over-Interpreting Negative Screens in Context of Withdrawal \newline 
- Description: A negative barbiturate screen during withdrawal is expected and does not rule out a history of dependence; it should not be used to infer current active use. \newline 
- Content: The Actor correctly noted the negative barbiturate screen but did not fully integrate this into the clinical picture\u2014this is normal in withdrawal, not a sign of resolution. The focus should remain on the history of dependence, not the current test result, which is not diagnostic of current state. \newline 
 \newline 
\#\# Use Temporal Context to Distinguish Acute from Chronic Conditions \newline 
- Description: Acute conditions must be supported by recent, active findings; past diagnoses with no current evidence should not be reasserted as current conditions. \newline 
- Content: The patient's past diagnosis of "drug withdrawal syndrome" (2174-12-30) and "barbiturate dependence" (2174-12-30) are outdated. The current stable vitals and lack of withdrawal symptoms suggest these are resolved, not active. The Actor failed to account for time since last event, leading to overestimation of current relevance. \newline 
 \newline 
\#\# Integrate Comorbidities with Core Diagnoses, Not as Standalone Entries \newline 
- Description: Alcohol use (elevated ethanol) and substance use are comorbidities, not independent diagnoses; they should be grouped under a single, broader condition when appropriate. \newline 
- Content: The elevated ethanol level (83 mg/dL) and history of barbiturate dependence point to a pattern of substance use, but the correct approach is to group them under a single overarching diagnosis like "Substance-related disorders" rather than creating multiple separate entries, especially when the system already has a unified category. \newline 
 \newline 
\#\# Leverage ICD-10 Code Mapping to Validate Diagnoses \newline 
- Description: When a diagnosis is suspected, cross-check the ICD-10 code to ensure it aligns with the most specific and correct category, not just a general one. \newline 
- Content: The ground truth includes "F10129" (Alcohol-related disorders) and "F17200" (Cocaine-related disorders), which are more specific than the generic "Substance-related disorders" (F1910). The Actor should have used the ICD-10 code mapping to identify the most precise diagnosis, not default to a broad category. \newline 
\end{case}

\begin{case}
\label{case: summarizer experience}
\textbf{Summarizer Experience} \newline 
\#\# Preserve exact ICD codes for precise diagnosis mapping \newline 
- Description: Failing to map ICD-9 codes to their correct ICD-10 equivalents can result in missing critical diagnoses. \newline 
- Content: The ground truth includes "Alcohol-related disorders" (F10129), which was not identified in the summary despite being a high-similarity match (0.580) for barbiturate dependence (30410). The summary incorrectly prioritized "Substance-related disorders" over the more specific "Alcohol-related disorders\ \newline  leading to a key diagnostic omission. Summarizers must map ICD-9 codes to their most accurate ICD-10 counterparts using official crosswalks, not just semantic similarity. \newline 
\newline 
\#\# Prioritize contextually relevant ICD-10 codes over general categories \newline 
- Description: General CCS categories like "Substance-related disorders" may mask more specific, clinically significant diagnoses. \newline 
- Content: The ground truth lists two distinct substance-related ICD-10 codes (F1910 and F17200) that are more specific than the generic "Substance-related disorders" used in the prediction. The summary failed to extract these specific codes, likely due to over-reliance on broad semantic matches. Summarizers should flag and include all high-similarity, specific ICD-10 candidates, especially when multiple variants exist. \newline 
\newline 
\#\# Capture comorbidities from past history with clinical context \newline 
- Description: Past diagnoses of substance use and withdrawal are not standalone; they often co-occur with mood disorders and alcohol-related conditions. \newline 
- Content: The patient\u2019s history of drug withdrawal (2920) and barbiturate dependence (30410) strongly suggests an underlying mood disorder, which is confirmed in the ground truth as "Mood disorders" (F339). The summary did not link this history to mood pathology, instead focusing only on current lab findings. Summarizers must infer and include comorbid conditions based on chronic substance use patterns. \newline 
\newline 
\#\# Use temporal context to validate current vs. past conditions \newline 
- Description: A recent positive tricyclic screen indicates current use, not just past history, and should be linked to active diagnosis. \newline 
- Content: The tricyclic antidepressant screen (15 mg/dL) was correctly noted as positive, but the summary treated it as a static finding without linking it to the current state. The ground truth includes "Substance-related disorders" (F17200) for barbiturate use, but the tricyclic use is a different class of psychotropic agent. The summary should have flagged this as a new, active condition requiring its own diagnostic code, not just a semantic match. \newline 
\newline 
\#\# Explicitly report abnormal values with units and reference ranges \newline 
- Description: Omitting or generalizing abnormal lab values can lead to misinterpretation of clinical status. \newline 
- Content: The ethanol level of 83 mg/dL was reported as "above normal" but not compared to the reference range (70$\sim$100 mg/dL), which is crucial for diagnosing alcohol use. The summary generalized this to "may indicate recent alcohol use" without stating the exact value and range. Summarizers must always preserve the numeric value, unit, and reference range to support accurate clinical inference. \newline 
\end{case}

\begin{case}
\label{case: trajectory}
\textbf{\agentname Reasoning Sample (Observation Cleaned)} \newline
\# \textbf{Task Information} \newline
- Task: Diagnoses \newline
- SubjectID: 19505049 \newline
- TimeStamp: 2174-04-07 16:53:00 \newline
- Label: [Secondary malignancies, Phlebitis; thrombophlebitis and thromboembolism, Fluid and electrolyte disorders, Secondary malignancies, Other gastrointestinal disorders, Coagulation and hemorrhagic disorders, Thyroid disorders, Mood disorder, Screening and history of mental health and substance abuse codes, Cardiac dysrhythmias, Residual codes; unclassified, Cancer of bronchus; lung, Residual codes; unclassified, Residual codes; unclassified, Phlebitis; thrombophlebitis and thromboembolism, Other aftercare, Cancer of other GI organs; peritoneum] \newline
 \newline
\# \textbf{Reasoning Trajectory} \newline
\#\# Step 1 \newline
- \textbf{Action}: think\{'response': 'EHR loaded. Plan: 1) Inspect admissions/patient/demographics and timeline. 2) Review diagnoses candidate table. 3) Extract clinical events (notes, labs, radiology, prescriptions). 4) Cross-reference with CCS candidates. 5) Synthesize plausible diagnoses list.'\} \newline
- \textbf{Observation}: Thinking Finish \newline
\newline
\#\# Step 2 \newline
- \textbf{Action}: get\_table\_description\{'table\_name': 'admissions'\} \newline
- \textbf{Observation}: Table: admissions (hosp). Contains details for each hospital admission (demographics, timing, source, discharge info). Key Columns: subject\_id, hadm\_id, admittime, dischtime, admission\_type, diagnosis, hospital\_expire\_flag. \newline
\newline
\textit{(Steps 3-9: Retrieved table descriptions for patients, omr, labevents, radiology, prescriptions, discharge - Omitted for brevity)} \newline
\newline
\#\# Step 10 \newline
- \textbf{Action}: Summary\{'response': 'Plan: Inspect admission details, vital signs, lab results, imaging, and discharge notes to build a diagnosis list.'\} \newline
- \textbf{Observation}: OK \newline
\newline
\#\# Step 11 \newline
- \textbf{Action}: get\_records\_by\_value\{'subject\_id': '19505049', 'table\_name': 'admissions', 'column\_name': 'subject\_id', 'value': '19505049'\} \newline
- \textbf{Observation}: Admission Details: \newline
  - Subject ID: 19505049 | Hadm ID: 21468882 \newline
  - Timeline: Admitted 2174-04-01 (Emergency) -> Discharged 2174-04-07 \newline
  - Demographics: Male, Married, White, English \newline
  - Chief Complaint: Abdominal pain \newline
  - History: Ampullary cancer s/p Whipple (18mo ago), liver mets. Presenting with worsening epigastric pain, weight loss (65lbs), and constipation. \newline
\newline
\#\# Step 12 \newline
- \textbf{Action}: get\_records\_by\_time\{'subject\_id': '19505049', 'table\_name': 'omr', 'start\_time': '2173-01-01', 'end\_time': '2174-04-07 16:53:00'\} \newline
- \textbf{Observation}: Outpatient Vitals Trend: \newline
  - 2173-04: BP 160/72, Wt 225.5 lbs \newline
  - 2173-09: BP 150/70, Wt 200 lbs \newline
  - 2174-03: BP 100/58, Wt 174 lbs (Significant weight loss noted) \newline
\newline
\#\# Step 13 \newline
- \textbf{Action}: get\_records\_by\_time\{'subject\_id': '19505049', 'table\_name': 'labevents', 'start\_time': '2174-04-01', 'end\_time': '2174-04-07 16:53:00'\} \newline
- \textbf{Observation}: Query returned 93,885 records. Key Abnormalities extracted in next step. \newline
\newline
\#\# Step 14 \newline
- \textbf{Action}: run\_sql\_query\{'subject\_id': '19505049', 'sql\_query': "SELECT label, value, valuenum, flag, charttime FROM labevents WHERE subject\_id='19505049' AND charttime BETWEEN '2174-04-01' AND '2174-04-07'..."\} \newline
- \textbf{Observation}: Key Lab Results (2174-04-03 to 04-04): \newline
  - Hematology: Hgb 8.9-9.0 L (Anemia), Hct 26.4-28.1 L, Plt 83-100 L (Thrombocytopenia). \newline
  - Chemistry: Na 127-130 L (Hyponatremia), K 5.4 H (hyperkalemia, normalized later), Glucose 95-115. \newline
  - Coagulation: INR 1.1, PTT 36.1 H. \newline
  - Renal: Creatinine 0.7-0.8 (Normal). \newline
\newline
\#\# Step 15 \newline
- \textbf{Action}: get\_records\_by\_time\{'subject\_id': '19505049', 'table\_name': 'radiology', 'start\_time': '2174-04-01', 'end\_time': '2174-04-07 16:53:00'\} \newline
- \textbf{Observation}: CT Abdomen w/ Contrast (2174-04-01): \newline
  - Impression: Increased size necrotic periportal mass (~4.5cm) with likely invasion into small bowel and encasement of SMA/SMV. \newline
  - Liver: Multiple hypodense lesions consistent with metastases (enlarged from prior). \newline
  - Vascular: Stable thrombus of the IVC. \newline
\newline
\#\# Step 16 \newline
- \textbf{Action}: get\_records\_by\_time\{'subject\_id': '19505049', 'table\_name': 'discharge', 'start\_time': '2174-04-07', 'end\_time': '2174-04-07 16:53:00'\} \newline
- \textbf{Observation}: Discharge Summary: \newline
  - Primary Diagnosis: Metastatic ampullary carcinoma, Abdominal Pain, Constipation. \newline
  - Secondary: IVC thrombosis, Chronic anemia, Hypothyroidism, A-fib, Depression. \newline
  - Hospital Course: Pain controlled with methadone. Anticoagulation (Coumadin/Heparin) discontinued due to GI bleed risk and thrombocytopenia. Palliative care consulted; patient referred to hospice. \newline
\newline
\#\# Step 17 \newline
- \textbf{Action}: get\_candidates\_by\_semantic\_similarity\{'query': ['ampullary carcinoma', 'ivc thrombosis', 'anemia', 'depression', ...], 'table\_name': 'diagnoses\_ccs\_candidates'\} \newline
- \textbf{Observation}: Top CCS Candidates Found: \newline
  - Cancer of pancreas (0.574), Secondary malignancies (0.702) \newline
  - Phlebitis; thrombophlebitis and thromboembolism (0.592) \newline
  - Deficiency and other anemia (0.845) \newline
  - Coagulation and hemorrhagic disorders (0.641) \newline
  - Thyroid disorders (0.775), Cardiac dysrhythmias (0.761), Mood disorders (0.647) \newline
\newline
\#\# Step 18 \newline
- \textbf{Action}: get\_records\_by\_time\{'subject\_id': '19505049', 'table\_name': 'prescriptions', 'start\_time': '2173-01-01', 'end\_time': '2174-04-07 16:53:00'\} \newline
- \textbf{Observation}: Active Medications: \newline
  - Pain/GI: Methadone, Docusate, Senna, Lactulose, Ranitidine. \newline
  - Cardiac: Diltiazem, Digoxin. \newline
  - Other: Levothyroxine, Vitamin D, Citalopram. \newline
  - Discontinued: Heparin (Infusion stopped 2174-04-03). \newline
\newline
\#\# Step 19 \newline
- \textbf{Action}: think\{'response': 'Synthesizing evidence: CT confirms metastatic mass and IVC thrombus. Labs show anemia/thrombocytopenia. History supports hypothyroidism/A-fib/Depression. Mapping findings to standard CCS categories.'\} \newline
- \textbf{Observation}: Thinking Finish \newline
\newline
\#\# Step 20 \newline
- \textbf{Action}: finish\{'response': ['Cancer of pancreas', 'Secondary malignancies', 'Phlebitis; thrombophlebitis and thromboembolism', 'Deficiency and other anemia', 'Coagulation and hemorrhagic disorders', 'Abdominal pain', 'Other gastrointestinal disorders', 'Thyroid disorders', 'Cardiac dysrhythmias', 'Mood disorders']\} \newline
- \textbf{Observation}: ['Cancer of pancreas', 'Secondary malignancies', 'Phlebitis; thrombophlebitis and thromboembolism', 'Deficiency and other anemia', 'Coagulation and hemorrhagic disorders', 'Abdominal pain', 'Other gastrointestinal disorders', 'Thyroid disorders', 'Cardiac dysrhythmias', 'Mood disorders']\newline
\newline
\# \textbf{Scores}\newline
- F1 Score: 0.608695652173913\newline
- Precision: 0.7\newline
- Recall: 0.5384615384615384\newline
\end{case}


\end{document}